\documentclass[10pt,twocolumn,letterpaper]{article}

\usepackage{cvpr}              %

\usepackage{graphicx}
\usepackage{amsmath}
\usepackage{amssymb}
\usepackage{booktabs}
\usepackage{pifont}%
\usepackage{enumitem}
\usepackage{dsfont}
\usepackage{minitoc}
\usepackage{multirow}
\usepackage[pagebackref,breaklinks,colorlinks]{hyperref}
\AtBeginDocument{%
  \addtolength\abovedisplayskip{-0.5\baselineskip}%
  \addtolength\belowdisplayskip{-0.5\baselineskip}%
}

\newcommand{\paptitle}{NoisyTwins}%

\usepackage[capitalize]{cleveref}
\crefname{section}{Sec.}{Secs.}
\Crefname{section}{Section}{Sections}
\Crefname{table}{Table}{Tables}
\crefname{table}{Tab.}{Tabs.}

\usepackage{mypreamble}

\def\cvprPaperID{9068} %
\def\confName{CVPR}
\def\confYear{2023}

\usepackage{breakurl}
\usepackage{adjustbox}
\usepackage[table,dvipsnames,svgnames,x11names]{xcolor}
\definecolor{darkgreen}{RGB}{30,150,30}

\begin{document}

\title{\paptitle: Class-Consistent and Diverse Image Generation through StyleGANs}

\author{Harsh Rangwani\textsuperscript{1}\thanks{Equal Contribution. Link: \href{https://rangwani-harsh.github.io/NoisyTwins/}{rangwani-harsh.github.io/NoisyTwins}} \footnotetext{Equal Contribution} \quad Lavish Bansal\textsuperscript{1,3}$^*$ \quad Kartik Sharma\textsuperscript{1,4} \quad Tejan Karmali\textsuperscript{2} \\ \quad Varun Jampani\textsuperscript{2}\quad R. Venkatesh Babu\textsuperscript{1} \\
\textsuperscript{1} Vision and AI Lab, IISc Bangalore \quad \textsuperscript{2} Google Research \quad \textsuperscript{3} IIT BHU Varanasi \quad \textsuperscript{4} BITS Pilani}
\maketitle

\begin{abstract}
StyleGANs are at the forefront of controllable image generation as they produce a latent space that is semantically disentangled, making it suitable for image editing and manipulation. However, the performance of StyleGANs severely degrades when trained via class-conditioning on large-scale long-tailed datasets.  We find that one reason for degradation is the collapse of latents for each class in the $\mathcal{W}$ latent space. With NoisyTwins, we first introduce an effective and inexpensive augmentation strategy for class embeddings, which then decorrelates the latents based on self-supervision in the  $\mathcal{W}$  space. This decorrelation mitigates collapse, ensuring that our method preserves intra-class diversity with class-consistency in image generation. We show the effectiveness of our approach on large-scale real-world long-tailed datasets of ImageNet-LT and iNaturalist 2019, where our method outperforms other methods by $\sim 19\%$ on FID, establishing a new state-of-the-art.
\end{abstract}

\section{Introduction}
StyleGANs~\cite{karras2019style, karras2020analyzing} have shown unprecedented success in image generation, particularly on well-curated and articulated datasets (eg. FFHQ for face images, etc.). In addition to generating high fidelity and diverse images, StyleGANs also produce a disentangled latent space, which is extensively used for image editing and manipulation tasks~\cite{wu2021stylespace}. As a result, StyleGANs are being extensively used in various applications like face-editing~\cite{shen2020interpreting, harkonen2020ganspace}, video generation~\cite{stylegan_v, digan}, face reenactment~\cite{bounareli2022finding}, etc., which are a testament to their usability and generality. 
However, despite being successful on well-curated datasets, training StyleGANs on in-the-wild and multi-category datasets is still challenging. A large-scale conditional StyleGAN (i.e. StyleGAN-XL) on ImageNet was recently trained successfully by Sauer \etal~\cite{Sauer2021ARXIV} using the ImageNet pre-trained model through the idea of a projection discriminator~\cite{Sauer2021NEURIPS}. While the StyleGAN-XL uses additional pre-trained models, obtaining such models for distinctive image domains like medical, forensics, and fine-grained data may not be feasible, which limits its generalization across domains.

\begin{figure}
    \centering
    \includegraphics[width=\columnwidth]{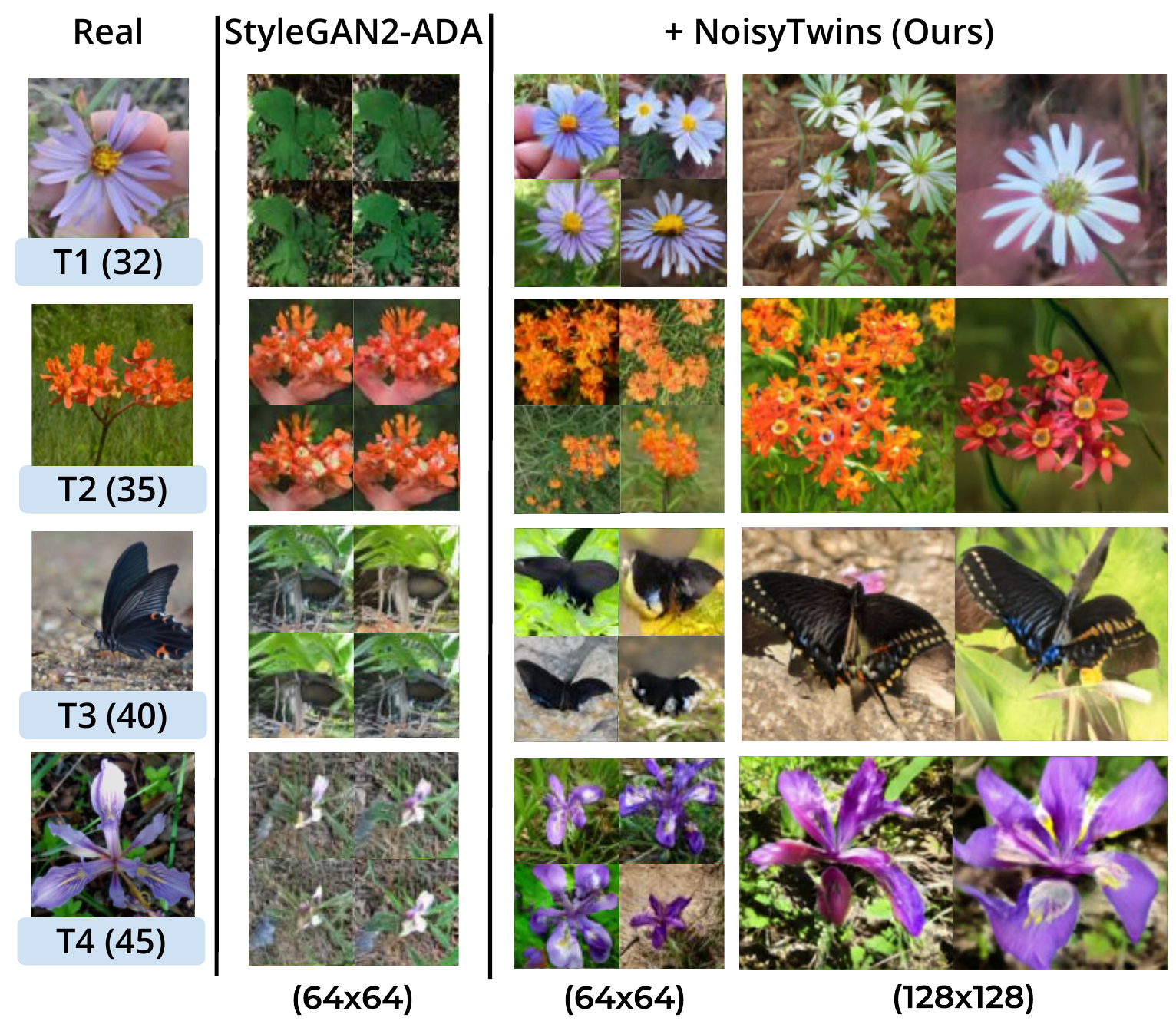}
    \caption{\textbf{Qualitative Comparison on tail classes (T1-T4) for iNaturalist 2019.} We provide sample(s) from real class (with class frequency), generated by StyleGAN2-ADA and after adding proposed NoisyTwins. NoisyTwins achieves remarkable diversity, class-consistency and quality by just using ~{38} samples on average.}
    \label{fig:result_teaser}
    \vspace{-5.0mm}
\end{figure}

\begin{figure*}[!t]
    \centering
    \includegraphics[width=\linewidth]{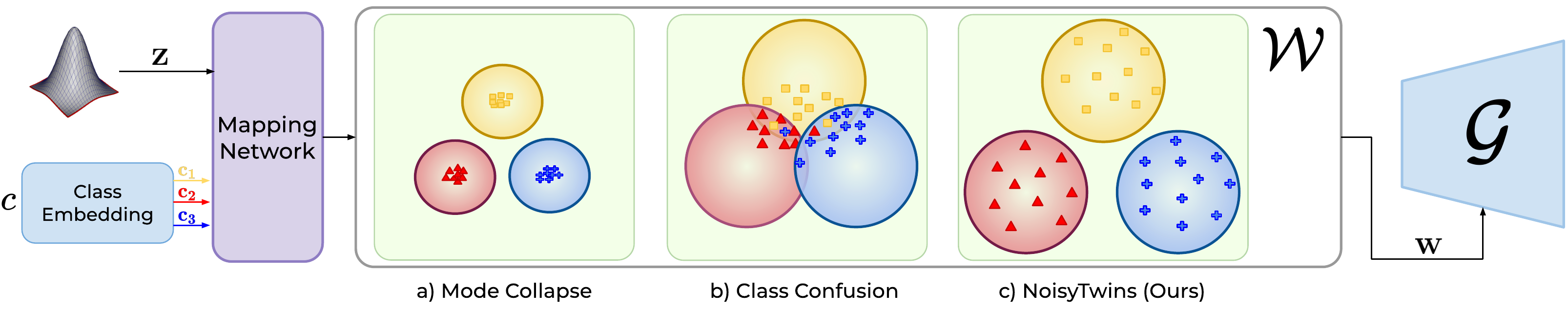}
    \caption{\textbf{Schematic illustration of $\mc{W}$ space for different GANs.} Existing conditioning methods either suffer from mode collapse~\cite{Karras2020ada} or lead to class confusion~\cite{rangwani2022gsr} in $\mc{W}$ space. With proposed NoisyTwins, we achieve intra class diversity while avoiding class confusion.}
    \label{fig:teaser}
     \vspace{-5mm}
\end{figure*}
In this work, we aim to train vanilla class-conditional StyleGAN without any pre-trained models on challenging real-world long-tailed data distributions. As training StyleGAN with augmentations~\cite{Karras2020ada, zhao2020diffaugment}  leads to low recall~\cite{improved_pr} (which measures diversity in the generated images) and mode collapse, particularly for minority (i.e. tail) classes.
For investigating this phenomenon further, we take a closer look at the latent $\mc{W}$ space of StyleGAN that is produced by a fully-connected mapping network that takes the conditioning variables $\mb{z}$ (i.e. random noise) and class embedding $\mb{c}$ as inputs. The vectors $\mb{w}$ in $\mc{W}$ space are used for conditioning various layers of the generator (Fig. \ref{fig:teaser}).  We find that output vectors $\mb{w}$ from the mapping network hinge on the conditioning variable $\mb{c}$ and become invariant to random conditioning vector $\mb{z}$. This collapse of latents leads to unstable training and is one of the causes of poor recall (a.k.a. mode collapse) for minority classes. Further, on augmenting StyleGAN with recent conditioning and regularization techniques~\cite{rangwani2022gsr, kang2021ReACGAN}, we find that they either lead to a poor recall for minority classes or lead to class confusion (Fig. \ref{fig:teaser}) instead of mitigating the collapse.

To mitigate the collapse of $\mb{w}$ in $\mc{W}$ space, we need to ensure that the change in conditioning variable $\mb{z}$ leads to the corresponding change in $\mb{w}$. Recently in self-supervised learning, several techniques~\cite{zbontar2021barlow, bardes2022vicreg} have been introduced to prevent the collapse of learned representations by maximizing the information content in the feature dimensions. Inspired by them we propose \paptitle{}, in which we first generate inexpensive twin augmentations for class embeddings and then use them to decorrelate the $\mb{w}$ variables through self-supervision. The decorrelation ensures that $\mb{w}$ vectors are diverse for each class and the GAN is able to produce intra-class diversity among the generated images.

We evaluate \paptitle{} on challenging benchmarks of large-scale long-tailed datasets of ImageNet-LT~\cite{liu2019large} and iNaturalist 2019~\cite{van2018inaturalist}. These benchmarks are particularly challenging due to a large number of classes present, which makes GANs prone to class confusion.
On the other hand, as these datasets are long-tailed with only a few images per class in tail classes, generating diverse images for those classes is challenging. %
We observe that existing metrics used in GAN evaluations are not able to capture both class confusion and mode collapse.
As a remedy, we propose to use intra-class Frechet Inception Distance (FID)~\cite{heusel2017gans} based on features obtained from pre-trained CLIP~\cite{radford2021learning} embeddings as an effective metric to measure the performance of class-conditional GANs in long-tailed data setups. 
Using \paptitle{} enables StyleGAN to generate diverse and class-consistent images across classes, mitigating the mode collapse and class confusion issues in existing state-of-the-art (SotA) (Fig.~\ref{fig:result_teaser}). Further, with NoisyTwins, we obtain diverse generations for tail classes even with $\leq$ 30 images, which can be attributed to the transfer of knowledge from head classes through shared parameters (Fig.~\ref{fig:result_teaser} and \ref{fig:qualitative_imgnet_lt}). In summary, we make the following contributions:

\begin{enumerate}[topsep=0pt,itemsep=-1ex,partopsep=1ex,parsep=1ex]
    \item We evaluate various recent SotA GAN conditioning and regularization techniques on the challenging task of long-tailed image generation. We find that all existing methods either suffer from mode collapse or lead to class confusion in generations.%
    \item To mitigate mode collapse and class confusion, we introduce \paptitle{}, an effective and inexpensive augmentation strategy for class embeddings that decorrelates latents in the $\mc{W}$ latent space (Sec.~\ref{sec:approach}).
    \item We evaluate NoisyTwins on large-scale long-tailed datasets of ImageNet-LT and iNaturalist-2019, where it consistently improves the StyleGAN2 performance ($\sim 19\%$), achieving a new SotA. Further, our approach can also prevent mode collapse and enhance the performance of few-shot GANs (Sec. \ref{subsec:few_shots_new}).
\end{enumerate}

\section{Related Works}
 \vspace{1mm}\noindent \textbf{StyleGANs.} Karras \etal introduced StyleGAN~\cite{karras2019style} and subsequently improved its image quality in StyleGAN2. StyleGAN could produce high-resolution photorealistic images as demonstrated on various category-specific datasets. It introduced a mapping network, which mapped the sampled noise into another latent space, which is more disentangled and semantically coherent, as demonstrated by its downstream usage for image editing and manipulation~\cite{Alaluf_2022_CVPR, Patashnik_2021_ICCV, shen2020interfacegan, shen2021closedform, parihar2022everything}. Further, StyleGAN has been extended to get novel views from images~\cite{FreeStyleGAN2021, liu2d3d, Shi2021Lifting2S}, thus making it possible to get 3D information from it. These downstream advances are possible due to the impressive performance of StyleGANs on class-specific datasets (such as faces). However, similar photorealism levels are yet uncommon on multi-class long-tailed datasets (such as ImageNet).

\vspace{1mm} \noindent  \textbf{GANs for Data Efficiency and Imbalance.}
Failure of GANs on less data was concurrently reported by Karras \etal ~\cite{Karras2020ada} and Zhao \etal ~\cite{zhao2020diffaugment}. The problem is rooted in the overfitting of the discriminator due to less real data. Since then, the proposed solutions for this problem have relied on a) augmenting the data, b) introducing regularizers, and c) architectural modifications. Karras \etal ~\cite{Karras2020ada} and Zhao \etal ~\cite{zhao2020diffaugment} relied on differentiable data augmentation before passing images into the discriminator to solve this problem. DeceiveD~\cite{jiang2021DeceiveD} proposed to introduce label-noise for discriminator training. LeCamGAN ~\cite{lecamgan} finds that enforcing LeCam divergence as a regularization trick in the discriminator can robustify GAN training under a limited data setting. DynamicD~\cite{yang2022improving} tunes the capacity of the discriminator on-the-fly during training. While these methods can handle the data inefficiency, they are ineffective on class imbalanced long-tailed data distribution~\cite{rangwani2022gsr}.

CBGAN~\cite{rangwani2021class} proposed a solution to train the unconditional GAN model on long-tailed data distribution by introducing a signal from the classifier to balance the classes generated by GAN. In a long-tailed class-conditional setting, gSR~\cite{rangwani2022gsr} proposes to regularize the exploding spectral norms of the class-specific parameters of the GAN. Collapse-by-conditioning~\cite{shahbazi2022collapse} addresses the limited data in classes by introducing a training regime that transitions from an unconditional to a class-conditioned setting, thus exploiting the shared information across classes during the early stages of the training. However, these methods suffer from either class confusion or poor generated image quality on large datasets, which is resolved by \paptitle{}.

\vspace{1mm} \noindent  \textbf{Self-Supervised Learning for GANs.} Ideas from Self-supervised learning have shown their benefits in GAN training. IC-GAN~\cite{casanova2021instanceconditioned} trains GAN conditioned on embeddings on SwAV~\cite{caron2020unsupervised}, which led to remarkable improvement in performance on the long-tailed version of ImageNet. InsGen~\cite{yang2021dataefficient} and ReACGAN~\cite{kang2020ContraGAN, kang2021ReACGAN} introduce the auxiliary task of instance discrimination for the discriminator, thereby making the discriminator focus on multiple tasks and thus alleviating discriminator overfitting. While InsGen relies on both noise space and image space augmentations, ReACGAN and ContraGAN follow only image space augmentations. Contrary to these, \paptitle{} performs augmentations in the class-embedding space and contrasts them in the $\mathcal{W}$-space of the generator instead of the discriminator.
\begin{figure*}[t]
    \centering
    \includegraphics[width=\linewidth]{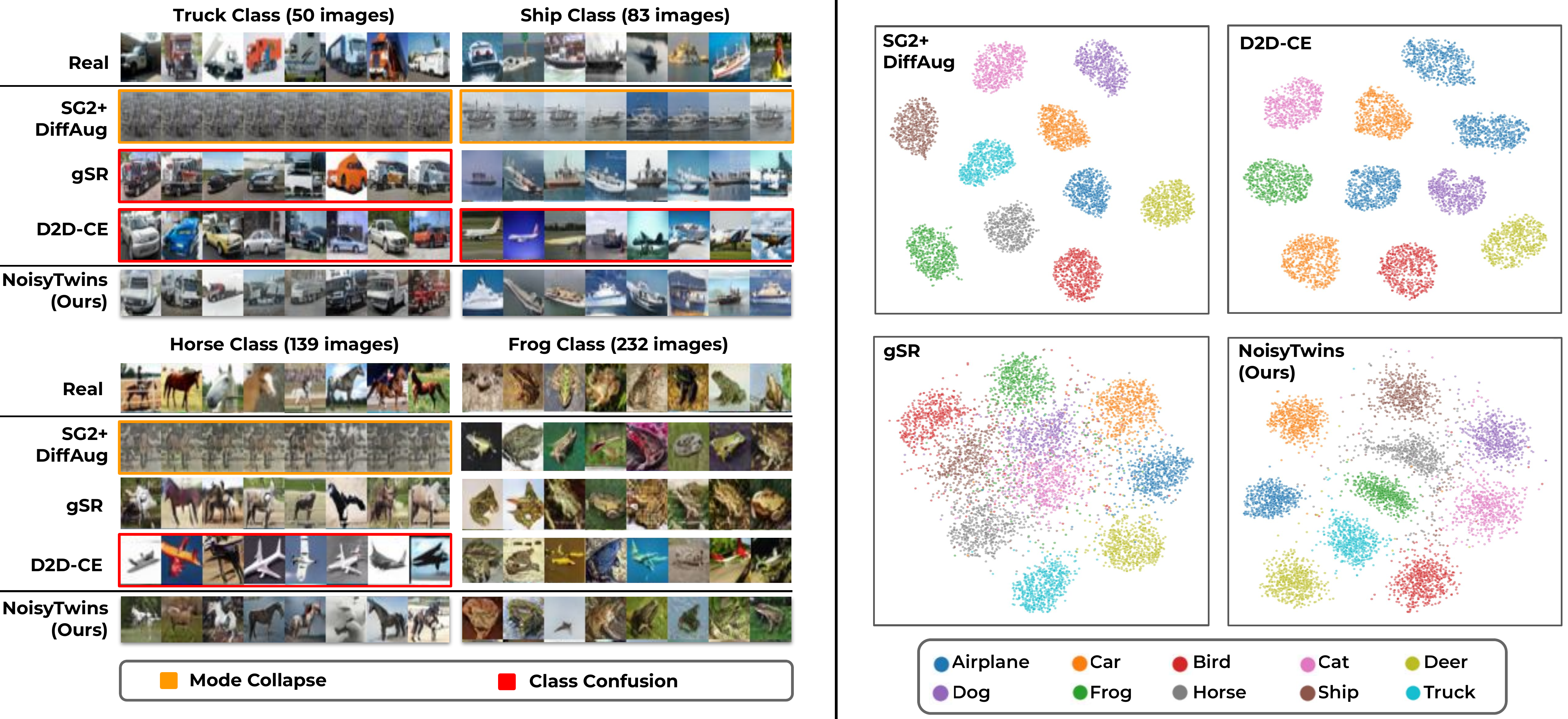}
     \caption{\textbf{Comparison of GANs and their $\mc{W}$ space for CIFAR10-LT.} We plot the generated images on (\textit{left}) and generate a t-SNE plot of $\mathbf{w}$ latents for generated images in $\mc{W}$ space (\textit{right}). We find that mode collapse and class confusion in images is linked to the corresponding collapse and confusion in latent $\mc{W}$ space. Our proposed NoisyTwins mitigates both collapse (\textit{left}) and confusion (\textit{right}) simultaneously.}
    \label{fig:cifar10_comparisons_motivation}
    \vspace{-5mm}
\end{figure*}

\vspace{-5mm}
\section{Preliminaries}

\subsection{StyleGAN}
StyleGAN~\cite{karras2019style} is a Generative Adversarial Network comprising of its unique Style Conditioning Based Generator ($\mc{G}$) and discriminator network ($\mc{D}$) trained jointly. We will focus on the architecture of StyleGAN2~\cite{Karras2019stylegan2} as we use it in our experiments, although our work is generally applicable to all StyleGAN architectures. The StyleGAN2 generator is composed of blocks that progressively upsample the features and resolution inspired by Progressive GAN~\cite{karras2017progressive}, starting from a single root image. The diversity in the images comes from conditioning each block of image generation through conditioning on the latent coming from the mapping network (Fig. \ref{fig:teaser}). The mapping network is a fully connected network that takes in the conditioning variables, the $\mb{z} \in \mbb{R}^d$ coming from a random distribution (e.g., Gaussian, etc.) and class conditioning label $c$ which is converted to an embedding $\mb{c} \in \mbb{R}^d$. The mapping network takes these and outputs vectors $\mb{w}$ in the $\mc{W}$ latent space of StyleGAN, which is found to be semantically disentangled to a high extent~\cite{wu2021stylespace}. The $\mb{w}$ is then processed through an affine transform and passed to each generator layer for conditioning the image generation process through Adaptive Instance Normalization (AdaIN)~\cite{huang2017arbitrary}. The images from generator $\mc{G}$, along with real images, are passed to discriminator $\mc{D}$ for training. The training utilizes the non-saturating adversarial losses~\cite{goodfellow2020generative} for $\mc{G}$ and $\mc{D}$ given as:
\begin{flalign}
    \min_{\mc{D}} \mc{L}_{\mc{D}} &= \sum_{i=1}^{m} \log(\mc{D}(\mb{x}_i)) + \log(1 - \mc{D}(\mc{G}(\mb{z}_i, \mb{c}_i))) \\
     \min_{\mc{G}} \mc{L}_{\mc{G}} &= \sum_{i=1}^{m} -\log(\mc{D}(\mc{G}(\mb{z}_i, \mb{c}_i)))
\end{flalign}
We now describe the issues present in the StyleGANs trained on long-tailed data and their analysis in $\mc{W}$ space.

\subsection{Class Confusion and Class-Specific Mode Collapse in Conditional StyleGANs}
To finely analyze the performance of StyleGAN and its variants on long-tailed datasets, we train them on the CIFAR10-LT dataset. In Fig. \ref{fig:cifar10_comparisons_motivation}, we plot the qualitative results of generated images and create a t-SNE plot for latents in $\mc{W}$ space for each class.  
We first train the StyleGAN2 baseline with augmentations (DiffAug)~\cite{Karras2020ada, zhao2020diffaugment}. We find that it leads to mode collapse, specifically for tail classes (Fig. \ref{fig:cifar10_comparisons_motivation}). In conjunction with images, we also observe that corresponding t-SNE embeddings are also collapsed near each class's mean in $\mc{W}$ space. Further, recent methods which have proposed the usage of contrastive learning for GANs, improve their data efficiency and prevent discriminator overfitting~\cite{kang2020ContraGAN, jeong2021contrad}. We also evaluate them by adding the contrastive conditioning method, which is D2D-CE loss-based on ReACGAN~\cite{kang2021ReACGAN}, to the baseline, where in results, we observe that the network omits to learn tail classes and produces head class images at their place (i.e., class confusion). In Fig.~\ref{fig:cifar10_comparisons_motivation}, it can be seen that the network confuses semantically similar classes, that is, generating cars (head or majority class) in place of trucks and airplanes (head class) instead of ships. In the $\mc{W}$ space, we find the same number of clusters as the number of classes in the dataset. However, the tail label cluster images also belong to the head classes of cars and airplanes. In a very recent work gSR~\cite{rangwani2022gsr}, it has been shown that constraining the spectral norm of $\mc{G}$ embedding parameters can help reduce the mode collapse and lead to stable training. However, we find that constraining the embeddings leads to class confusion, as seen in t-SNE visualization in Fig.~\ref{fig:cifar10_comparisons_motivation}. We find that this class confusion gets further aggravated when StyleGAN is trained on datasets like ImageNet-LT, which contain a large number of classes, along with a bunch of semantically similar classes (Sec.~\ref{subsec:result_inat_imnet}).  
Based on our $\mc{W}$ space analysis and qualitative results above, we observe that the class confusion and mode collapse of images is tightly coupled with the structure of $\mc{W}$ space. Further, the recent SotA methods are either unable to prevent collapse or suffer from class confusion. Hence, this work aims to develop a technique that mitigates both confusion and collapse.

\section{Approach}
\label{sec:approach}
In this section, we present our method \paptitle{}, which introduces noise-based augmentation twins in the conditional embedding space (Sec. \ref{subsec: noise-aug}), and then combines it with the Barlow-Twins-based regularizer from the self-supervised learning (SSL) paradigm to resolve the issue of class confusion and mode collapse (Sec.~\ref{subsec:bt}).

\subsection{Noise Augmentation in Embedding Space}
\label{subsec: noise-aug}
As we observed in the previous section that $\mb{w}$ vectors for each sample become insensitive to changes in $\mb{z}$. This collapse in $\mb{w}$ vectors for each class leads to mode collapse for baselines (Fig. \ref{fig:cifar10_comparisons_motivation}). One reason for this could be the fact that $\mb{z}$ is composed of continuous variables, whereas the embedding vectors $\mb{c}$ for each class are discrete.
Due to this, the GAN converges to the easy degenerate solution where it generates a single sample for each class, becoming insensitive to changes in $\mb{z}$. For inducing some continuity in $\mb{c}$ embeddings vectors, we introduce an augmentation strategy where we add i.i.d. noise of small magnitude in each of the variables in $\mb{c}$. Based on our observation (Fig.~\ref{fig:cifar10_comparisons_motivation}) and existing works~\cite{rangwani2022gsr}, there is a high tendency for mode collapse in tail classes. Hence we add noise in embedding space that is proportional to the inverse of the frequency of samples . We provide the mathematical expression of noise augmentation $\mb{\tilde{c}}$ below:
\begin{equation}
    \mb{\tilde{c}} \sim \mb{c} + \mc{N}(\mathbf{0}, \sigma_c\mbb{I}_d) \; \text{where} \; \sigma_c = \sigma \frac{( 1 - \alpha)}{1 - \alpha^{n_{c}}} 
    \label{eq:noise_aug_eq}
\end{equation}
Here $n_c$ is the frequency of training samples in class $c$, $\mbb{I}$ is the identity matrix of size $d \times d$, and $\alpha, \sigma$ are hyper-parameters. The expression of $\sigma_c$ is from the effective number of samples~\cite{cui2019class}, which is a softer version of inverse frequency proportionality. In contrast to the image space augmentation, these \emph{noise augmentations come for free} as there is no significant additional computation overhead. 
This noise is added in the embedding $\mb{c}$ before passing it to the generator and the discriminator, which ensures that the class embeddings occupy a continuous region in latent space. \\ 
\begin{figure*}[t]
    \centering
    \includegraphics[width=\linewidth]{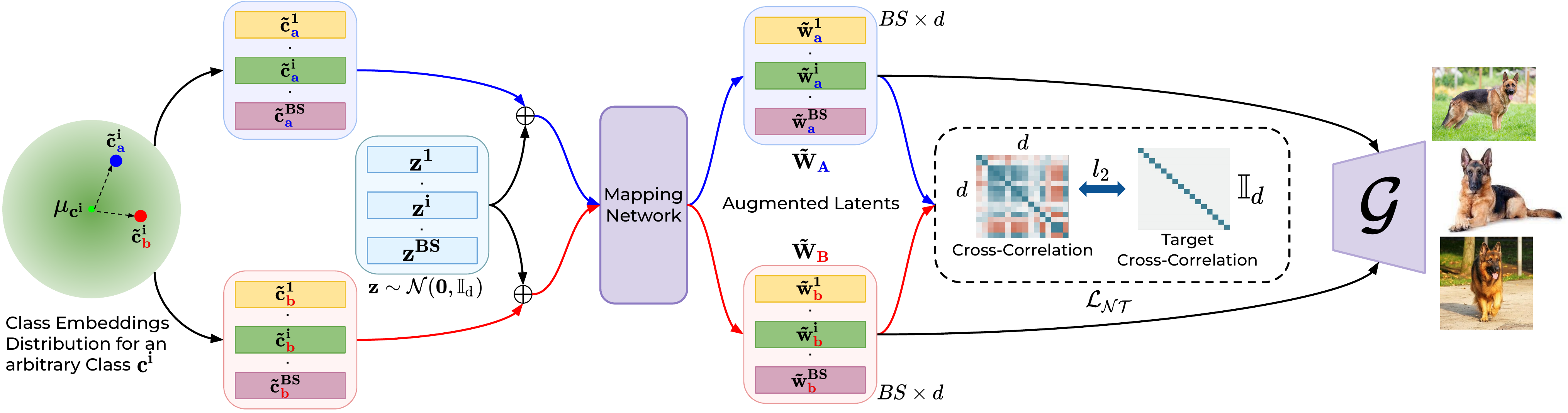}
    \caption{\textbf{Overview of NoisyTwins.} For the $i^{th}$ sample of class $c^i$, we create twin augmentations ($\tilde{\mb{c}}^{i}_a$, $\tilde{\mb{c}}^{i}_b$), by sampling from a Gaussian centered at class embedding ($\mu_{\mb{c{^i}}}$).  After this, we concatenate them with the same $\mb{z}^{i}$ and obtain ($\tilde{\mb{w}}^{i}_{a}, \tilde{\mb{w}}^{i}_{b}$) from the mapping network, which we stack in batches of augmented latents ($\mb{\tilde{W}_{A}}$ and $\mb{\tilde{W}_{B}}$). The twin ($\tilde{\mb{w}}^{i}_{a}, \tilde{\mb{w}}^{i}_{b}$) vectors are then made invariant to augmentations (similar) in the latent space by minimizing cross-correlation~\cite{bardes2022vicreg, zbontar2021barlow} between the latents of two augmented batches ($\mb{\tilde{W}_{A}}$ and $\mb{\tilde{W}_{B}}$). }
    \label{fig:overview_noisy}
    \vspace{-5mm}
\end{figure*}
\noindent\textbf{Insight:} The augmentation equation above (Eq. \ref{eq:noise_aug_eq})  can be interpreted as approximating the discrete random variable $\mb{c}$ with a Gaussian with finite variance and the embedding parameters $\mb{c}$ being the mean $\mb{\mu_{c}}$. 
\begin{equation}
    \mb{\tilde{c}} \sim \mc{N}(\mb{\mu_{c}}, \sigma_c \mbb{I}_d)
\end{equation}
This leads to the class-embedding input $\mb{\tilde{c}}$ to mapping network to have a Gaussian distribution, similar in nature to $\mb{z}$. This noise augmentation strategy alone mitigates the degenerate solution of class-wise mode collapse to a great extent (Table~\ref{tab:imageNetLT_iNat}) and helps generate diverse latent $\mb{w}$ for each class. Due to the diverse $\mb{w}$ conditioning of the GAN, it leads to diverse image generation. 

\subsection{Invariance in \texorpdfstring{$\mc{W}$}{W}-Space  with \paptitle{}}
\label{subsec:bt}
The augmentation strategy introduced in the previous section expands the region for each class in $\mc{W}$ latent space. Although that does lead to diverse image generation as $\mb{w}$ are diverse; however, this does not ensure that these $\mb{w}$ will generate class-consistent outputs for augmentations in embedding ($\tilde{\mb{c}}$). To ensure class consistent predictions, we need to ensure invariance in $\mb{w}$ to noise augmentation.

For enforcing invariance to augmentations, a set of recent works~\cite{zbontar2021barlow, bardes2022vicreg, grill2020bootstrap} in self-supervised learning make the representations of augmentations similar through regularization. Among them, we focus on Barlow Twins as it does not require a large batch size of samples. Inspired by Barlow twins, we introduce NoisyTwins (Fig. \ref{fig:overview_noisy}), where we generate twin augmentations $\tilde{\mb{c}}_a$
and $\tilde{\mb{c}}_b$ of the same class embedding ($\mu_{c}$) and concatenate them to same $\mb{z}$. After creating a batch of such inputs, they are passed to the mapping network to get batches of augmented latents ($\tilde{\mb{W}}_A$ and $\tilde{\mb{W}}_B$). These batches are then used to calculate the cross-correlation matrix of latent variables given as: 
\begin{equation}
     \mb{C}_{j,k} = \frac{\underset{(\tilde{\mb{w}}_{a}, \tilde{\mb{w}}_{b}) \in (\tilde{\mb{W}}_{A}, \tilde{\mb{W}}_B)}{\sum} \tilde{\mb{w}}_{a}^{j} \tilde{\mb{w}}_{b}^{k}}{\underset{\tilde{\mb{w}}_{a} \in \tilde{\mb{W}}_{A} }{\sum} {\tilde{\mb{w}}_{a}^{j} \tilde{\mb{w}}_{a}^{j}} \underset{\mb{\tilde{w}}_{b} \in \tilde{\mb{W}}_{B} }{\sum} \tilde{\mb{w}}_{b}^{k} \tilde{\mb{w}}_{b}^{k}} 
\end{equation}
The  matrix $\mb{C}$ is a square matrix of size same as of latents $\mb{w}$. The final loss based on confusion matrix
is given as:
\begin{equation}
\label{eq:noisytwins}
    \mc{L}_{\mc{NT}} = \sum_{j} (1 - \mb{C}_{jj}^{2}) + \gamma\sum_{j\neq k} \mb{C}_{j,k}^2
\end{equation}
The first term tries to make the two latents $(\tilde{\mb{w}}_{a}$ and $\tilde{\mb{w}}_{b}$) invariant to the noise augmentation applied (i.e. similar), whereas the second term tries to de-correlate the different variables, thus maximizing the information in $\mb{w}$ 
 vector~\cite{zbontar2021barlow} (See Appendix). The $\gamma$ is the hyper-parameter that determines the relative importance of the two terms. This loss is then added to the generator loss term ($\mc{L_G} + \lambda L_{\mc{NT}}$) and optimized through backpropagation. The above procedure comprises our proposed method, \paptitle{} (Fig. \ref{fig:overview_noisy}), which we empirically evaluate in the subsequent sections.

\section{Experimental Evaluation}

\subsection{Setup}
\vspace{1mm} \noindent  \textbf{Datasets:} We primarily apply all methods on long-tailed datasets, as GANs trained on them are more prone to class confusion and mode collapse. We first report on the commonly used CIFAR10-LT dataset with an imbalance factor (\ie ratio of most to least frequent class) of 100. To show our approach's scalability and real-world application, we test our method on the challenging ImageNet-LT and iNaturalist 2019 datasets. The ImageNet-LT~\cite{liu2019large} is a long-tailed variant of the 1000 class ImageNet dataset, with a plethora of semantically similar classes (e.g., Dogs, Birds etc.), making it challenging to avoid class confusion. iNaturalist-2019~\cite{van2018inaturalist} is a real-world long-tailed dataset composed of 1010 different variants of species, some of which have fine-grained differences in their appearance. For such fine-grained datasets, ImageNet pre-trained discriminators~\cite{Sauer2021NEURIPS, Sauer2021ARXIV} may not be useful, as augmentations used to train the model makes it invariant to fine-grained changes.

\vspace{1mm} \noindent\textbf{Training Configuration:} We use StyleGAN2 architecture for all our experiments. All our experiments are performed using PyTorch-StudioGAN implemented by Kang \etal~\cite{kang2022StudioGAN}, which serves as a base for our framework. 
We use Path Length regularization (PLR) as it ensures changes in $\mc{W}$ space lead to changes in images, using a delayed PLR on ImageNet-LT following~\cite{Sauer2021ARXIV}. We use a batch size of 128 for all our experiments, with one G step per D step. Unless stated explicitly, we used the general training setup for StyleGANs from~\cite{kang2022StudioGAN}, including methods like $R_1$ regularization~\cite{mescheder2018training}, etc. More details on the exact training configuration for each dataset are provided in Appendix.

\begin{figure}[!t]
    \centering
    \includegraphics[width=\linewidth]{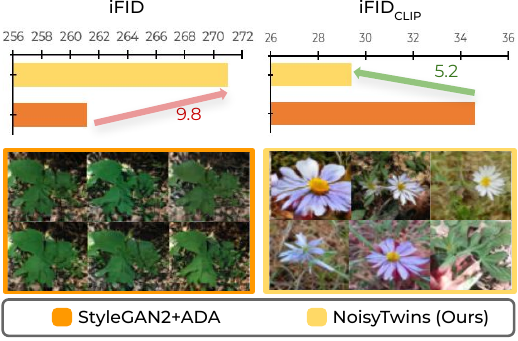}
    \caption{\textbf{Choice of Eval. backbone:} intra-FID (iFID) of a class based on InceptionV3 backbone (left plot) is not able to capture the mode collapse (increase in iFID in the absence of mode collapse). This is well-captured by iFID$_\mathrm{CLIP}$ based on CLIP~\cite{radford2021learning} backbone (right plot, decrease in iFID in the absence of mode collapse).}
    \label{fig:iFid Comparison}
    \vspace{-5mm}
\end{figure}

\vspace{1mm} \noindent \textbf{Metrics:}
In this work we use the following metrics for evaluation of our methods: \\
\noindent \textbf{a) FID:} Fr\'echet Inception Distance~\cite{heusel2017gans} is the Wasserstein-2 Distance between the real and validation data. We use a held-out validation set and 50k generated samples to evaluate FID in each case. As FID is biased towards ImageNet and can be arbitrarily manipulated,
we also report FID$_{\mathrm{CLIP}}$.\\ 
\noindent \textbf{b) Precision \& Recall:} As we aim to mitigate the mode collapse and achieve diverse generations across classes, we use improved Precision \& Recall~\cite{kynkaanniemi2019improved} metrics, as poor recall indicates mode collapse~\cite{Sauer2021ARXIV}. \\
\noindent \textbf{d) Intra-Class FID$_\mathrm{CLIP}$ (iFID$_\mathrm{CLIP}$):}
The usage of only FID based on Inception-V3 Networks for evaluation of Generative Models has severe limitations, as it has been found that FID can be reduced easily by some fringe features~\cite{Kynkaanniemi2022}. iFID is computed by taking FID between 5k generated and real samples for the same class. As we want to evaluate both class consistency and diversity, we find that similar limitations exist for intra-class FID (iFID), which has been used to evaluate class-conditional GANs~\cite{kang2022StudioGAN}.  In Fig.~\ref{fig:iFid Comparison}, we show the existence of generated images for a particular class (more in Appendix) from models trained on iNaturalist 2019, where iFID is better for the mode collapsed model than the other model generating diverse images. Whereas the iFID$_\mathrm{CLIP}$, based on CLIP backbone can rank the models correctly with the model having mode collapse having high iFID$_\mathrm{CLIP}$. Further, we find that the mean iFID can be deceptive in detecting class confusion and collapse cases, as it sometimes ranks models with high realism better than models generating diversity (See Appendix). Hence, mean iFID$_\mathrm{CLIP}$ (ref. to as iFID$_\mathrm{CLIP}$ in result section for brevity) can be reliably used to evaluate models for class consistency and diversity.

\begin{table*}[!t]
    \centering
    \parbox{\textwidth}{
    \caption{\textbf{Quantitative results on ImageNet-LT and iNaturalist 2019 Datasets.} We compare FID($\downarrow$), FID$_\mathrm{CLIP}$($\downarrow$), iFID$_\mathrm{CLIP}$($\downarrow$), Precision($\uparrow$) and Recall($\uparrow$) with other existing \mbox{approaches on StyleGAN2 (SG2).} We obtain an average $\sim19\%$ relative improvement on FID, $\sim33\%$ on FID$_\mathrm{CLIP}$, and $\sim11\%$ on iFID$_\mathrm{CLIP}$ metrics over the previous SotA on ImageNet-LT and iNaturalist 2019 datasets.
    }
    \vspace{-3mm}
        \label{tab:imageNetLT_iNat}}
    \resizebox{\textwidth}{!}{
    \begin{tabular}{lccccc|ccccc}
    \toprule
         & \multicolumn{5}{c}{ImageNet-LT} & \multicolumn{5}{c}{iNaturalist 2019} \\ \hline
         Method & FID($\downarrow$) & FID$_\mathrm{CLIP}$($\downarrow$) & iFID$_\mathrm{CLIP}$($\downarrow$) & Precision($\uparrow$) & Recall($\uparrow$) & FID($\downarrow$) & FID$_\mathrm{CLIP}$($\downarrow$) & iFID$_\mathrm{CLIP}$($\downarrow$) & Precision($\uparrow$) & Recall($\uparrow$) \\
         \midrule
         SG2~\cite{Karras2019stylegan2}& 41.25 & 11.64 & 46.93 & 0.50 & \underline{0.48} & 19.34 & 3.33 & 38.24 & 0.74 & 0.17\\
         SG2+ADA~\cite{Karras2020ada}& 37.20 & 11.04 & 47.41 & 0.54 & 0.38 & 14.92 & 2.30 & 35.19 & 0.75 & 0.57\\
         SG2+ADA+gSR~\cite{rangwani2022gsr} & 24.78 & 8.21 & 44.42 & 0.63 & 0.35 & 15.17 & 2.06 & 36.22 & 0.74 & 0.46 \\
         \midrule
         
        \rowcolor{gray!10}  SG2+ADA+Noise (Ours) & \underline{22.17} & \underline{7.11} & \underline{41.20} & \textbf{0.72} & 0.33 & \underline{12.87} & \underline{1.37} & \textbf{31.43} & \textbf{0.81} & \underline{0.63}\\
        \rowcolor{gray!10} \; + NoisyTwins (Ours) & \textbf{21.29} & \textbf{6.41} & \textbf{39.74} & \underline{0.67} & \textbf{0.49} & \textbf{11.46} & \textbf{1.14} & \underline{31.50} & \underline{0.79} & \textbf{0.67}\\ \bottomrule
    \end{tabular}
    }
    \vspace{-2.00mm}
\end{table*}

\begin{table*}[t]
\parbox{.6\linewidth}{
\centering
    \caption{\textbf{Quantitative results on CIFAR10-LT Dataset.} We compare with other existing \mbox{approaches.} We obtain $\sim26\%$ relative improvement over the existing methods on FID$_\mathrm{CLIP}$ and iFID$_\mathrm{CLIP}$ metrics.}
    \label{tab:CIFAR10LT}
    \resizebox{0.65\textwidth}{!}
    {
    \begin{tabular}{lccccc}
    \toprule
         Method & FID($\downarrow$) & FID$_\mathrm{CLIP}$($\downarrow$) & iFID$_\mathrm{CLIP}$($\downarrow$) & Precision($\uparrow$) & Recall($\uparrow$) \\ \midrule
         SG2+DiffAug~\cite{zhao2020diffaugment}& 31.73 & 6.27 & 11.59 & 0.63 & 0.35\\
         SG2+D2D-CE~\cite{kang2021ReACGAN} & 19.97 & 4.77 & 11.35 & \textbf{0.73} & 0.42 \\ 
         gSR~\cite{rangwani2022gsr} & 22.10 & 5.54 & 9.94 & 0.70 & 0.29 \\
         \midrule
         
        \rowcolor{gray!10}  SG2+DiffAug+Noise (Ours) & 28.90 & 5.26 & 10.65 & 0.71 & 0.38\\
        \rowcolor{gray!10} \; + NoisyTwins(Ours) & \textbf{17.74} & \textbf{3.55} & \textbf{7.24} & 0.70 & \textbf{0.51}\\  \bottomrule
    \end{tabular}
    }
}
\hfill
\parbox{.35\linewidth}{
        \centering
    \caption{\textbf{Comparison with SotA approaches on BigGAN.} We compare FID($\downarrow$)  with other existing \mbox{models} on ImageNet-LT (IN-LT) and iNaturalist 2019 (iNat-19).}
    \label{tab:SotA_compare}
    \vspace{-4mm}
    \resizebox{0.3\textwidth}{!}
    {\begin{tabular}{lcc}
    \toprule
         Method &  iNat-19  & IN-LT \\ \midrule
         BigGAN~\cite{brock2018large} & 14.85 & 28.10\\
         \; + gSR~\cite{rangwani2022gsr} & 13.95 & -\\
         ICGAN~\cite{casanova2021instanceconditioned}  & - & 23.40\\
         \midrule
         
        \rowcolor{gray!10}  StyleGAN2-ADA~\cite{Karras2020ada} &  14.92 & 37.20\\
        \rowcolor{gray!10}  + NoisyTwins (Ours) & \textbf{11.46} & \textbf{21.29}\\ \bottomrule
    \end{tabular}}
}
\vspace{-4mm}
\end{table*}

\noindent \textbf{Baselines:}
For evaluating NoisyTwins performance in comparison to other methods, we use the implementations present in StudioGAN~\cite{kang2020ContraGAN}. For fairness, we re-run all the baselines on StyleGAN2 in the same hyperparameter setting. We compare our method to the StyleGAN2 (SG2)
and StyleGAN2 with augmentation (DiffAug~\cite{zhao2020diffaugment} and ADA~\cite{Karras2020ada}) baselines. We further tried to improve the baseline by incorporating the recent LeCam regularization method; however, it resulted in gains only for the iNaturalist 2019 dataset, where we use LeCam for all experiments. Further on StyleGAN2, we also use contrastive D2D-CE loss conditioning (\ie ReACGAN) as a baseline.   However, the D2D-CE baseline completely ignores learning of tail classes (Fig. \ref{fig:cifar10_comparisons_motivation}) for CIFAR10-LT and is expensive to train; hence we do not report results for it for large-scale long-tailed datasets. We also compare our method against the recent SotA group Spectral Normalization (gSR)~\cite{rangwani2022gsr} method, which we implement for StyleGAN2 by constraining the spectral norms of embedding parameters of the generator ($\mc{G}$) as suggested by authors. As a sanity check, we reproduce their results on CIFAR10-LT and find that our implementation matches the reported results correctly. We provide results on all datasets, for both the proposed Noise Augmentation (+ Noise) and the overall proposed NoisyTwins (+NoisyTwins) method.

\subsection{Results on Long-Tailed Data Distributions}
\label{subsec:result_inat_imnet}

\begin{figure*}[t]
    \centering
    \includegraphics[width=\linewidth]{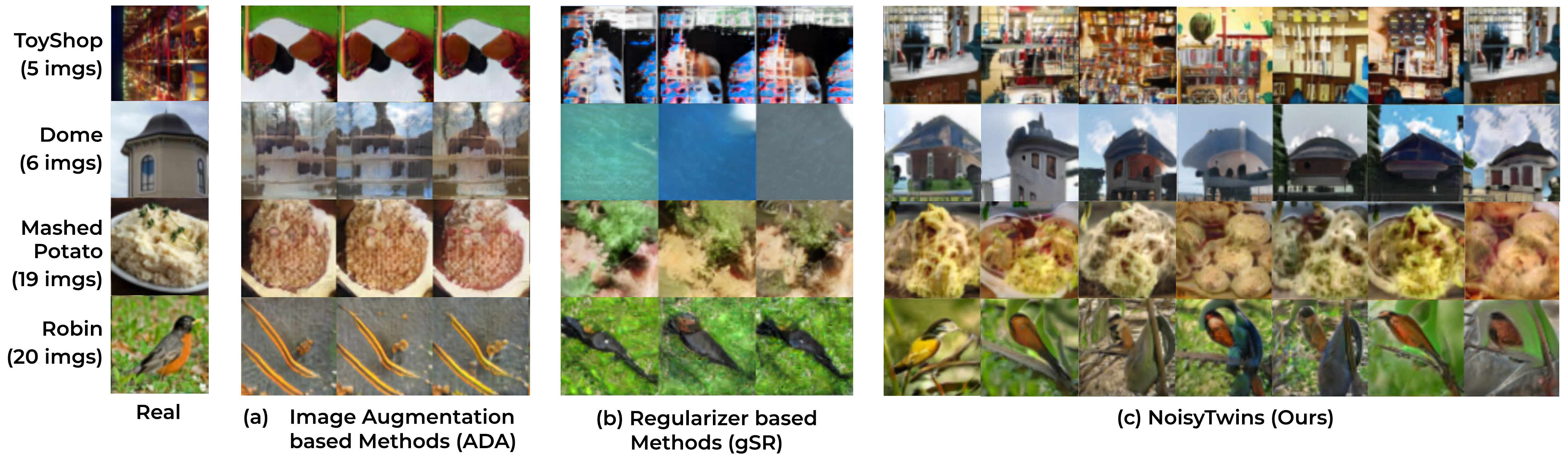}
    \caption{\textbf{Qualitative results on ImageNet-LT for tail classes.} We find that existing SotA methods for tail classes show collapsed (a) or arbitrary image generation (b). With NoisyTwins, we observe diverse and class-consistent image generation, even for classes having 5-6 images. The tail classes get enhanced diversity by transferring the knowledge from head classes, as they share parameters.}
    \label{fig:qualitative_imgnet_lt}
    \vspace{-5mm}
\end{figure*}

\noindent\textbf{CIFAR10-LT}. We applied DiffAug~\cite{zhao2020diffaugment} on all baselines, except on gSR, where we found that DiffAug provides inferior results compared to ADA (as also used by authors~\cite{rangwani2022gsr}). It can be observed in Table~\ref{tab:CIFAR10LT} that the addition of NoisyTwins regularization significantly improves over baseline by ($\sim$ 14 FID) along with providing superior class consistency as shown by improved iFID$_\mathrm{CLIP}$. NoisyTwins is also able to outperform the recent gSR regularization method and achieves improved results for all metrics. Further, NoisyTwins improves FID for StyleGAN2-ADA baseline used by  gSR too from 32.08 to 23.02, however the final results are inferior than reported DiffAug baseline results. Further, we observed that despite not producing any tail class images (Fig.~\ref{fig:cifar10_comparisons_motivation}), the D2D-CE baseline has much superior FID in comparison to baselines. Whereas the proposed iFID$_\mathrm{CLIP}$ value is similar for the baseline and D2D-CE model. This clearly demonstrates the superiority of proposed iFID$_\mathrm{CLIP}$ in detecting class confusion. %

\noindent\textbf{Large-scale Long-Tailed Datasets.} We experiment with iNaturalist 2019 and ImageNet-LT. These datasets are particularly challenging as they contain long-tailed imbalances and semantically similar classes, making GANs prone to mode collapse and class confusion. The baselines StyleGAN2 and StyleGAN2-ADA both suffer from mode collapse (Fig. \ref{fig:qualitative_imgnet_lt}), particularly for the tail classes.  Whereas for the recent SotA gSR method, we find that although it undergoes less collapse in comparison to baselines, it suffers from class confusion as seen from similar Intra-FID$_\mathrm{CLIP}$ in comparison to baselines (Table~\ref{tab:imageNetLT_iNat}). Compared to that, our method NoisyTwins improves when used with StyleGAN2-ADA significantly, leading to a relative improvement of $42.7\%$ in FID for ImageNet-LT and  $23.19\%$ on the iNaturalist 2019 dataset when added to StyleGAN2-ADA baseline. Further with Noise Augmentation (+Noise), we observe generations of high-quality class-consistent images, but it also suffers from mode collapse. This can be observed by the high-precision values in comparison to low-recall values. However, adding NoisyTwins regularization over the noise augmentation improves diversity by improving recall (Table \ref{tab:imageNetLT_iNat}). 

Fig. \ref{fig:qualitative_imgnet_lt} presents the generated images of tail classes for various methods on ImageNet-LT , where NoisyTwins generations show remarkable diversity in comparison to others. The presence of diversity for classes with just 5-6 training images demonstrates successful transfer of knowledge from head classes to tail classes, due to shared parameters. Further, to compare with existing SotA reported results, we compare FID of BigGAN models from gSR~\cite{rangwani2022gsr} and Instance Conditioned GAN (ICGAN)~\cite{casanova2021instanceconditioned}. For fairness, we compare FID on the validation set for which we obtained gSR models from authors and re-evaluate them, as they reported FID on a balanced training set. As  BigGAN models are more common for class-conditioned generation~\cite{kang2022StudioGAN}, their baseline performs superior to StyleGAN2-ADA baselines (Table~\ref{tab:SotA_compare}). However, the addition of NoisyTwins to the StyleGAN2-ADA method improves it significantly, even outperforming the existing methods of gSR (by 18.44\%) and ICGAN (by 9.44\%)  based on BigGAN architecture. This shows that NoisyTwins allows the StyleGAN2 baseline to scale to large and diverse long-tailed datasets.
\subsection{NoisyTwins on Few-Shot Datasets}
\label{subsec:few_shots_new}
We now demonstrate the potential of NoisyTwins in another challenging scenario of class-conditional few-shot image generation from GANs. We perform our experiments using a conditional StyleGAN2-ADA baseline, for which we tune hyper-parameters to obtain a strong baseline. We then apply our method of Noise Augmentation and NoisyTwins over the strong baseline for reporting our results. We use the few-shot dataset of LHI-AnimalFaces~\cite{si2011learning} and a subset of ImageNet Carnivores~\cite{liu2019few, shahbazi2022collapse} to report our results. Table~\ref{tab:few_shot} shows the results of these experiments, where we find that our method, NoisyTwins, significantly improves the FID of StyleGAN2 ADA baseline by (22.2\%) on average for both datasets. Further, combining Noisy Twins with SotA Transitional-cGAN~\cite{shahbazi2022collapse} through official code, also leads to effective improvement in FID. These results clearly demonstrate the diverse potential and applicability of our proposed method NoisyTwins.
\section{Analysis}
We perform analysis of NoisyTwins w.r.t. to its hyperparameters, standard deviation ($\sigma$) of noise augmentation and regularization strength ($\lambda$). We also compare NoisyTwins objective (ref. Eq.~\ref{eq:noisytwins}) with contrastive objective. Finally, we compare NoisyTwins over Latent Diffusion Models for long-tailed class conditional generation task. We perform ablation experiments on CIFAR10-LT, for which additional details and results are present in Appendix. We also present comparison of NoisyTwins for GAN fine-tuning. 

\begin{figure}[!t]
    \centering
    \includegraphics[width=\linewidth]{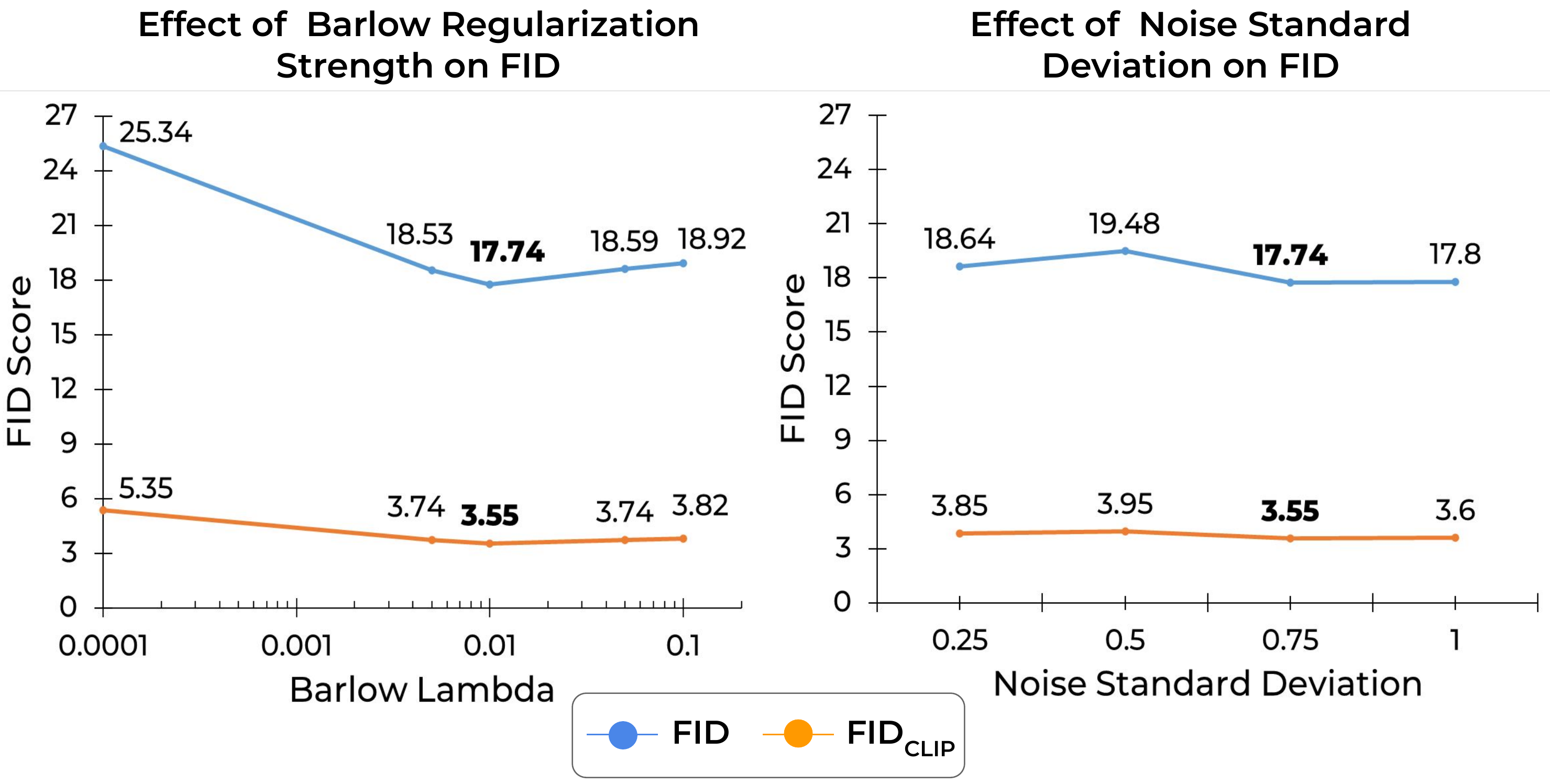}
    \caption{\textbf{Ablation of Hyperparameters.} Quantitative comparison on CIFAR10-LT for standard deviation of Noise Augmentation ($\sigma$) and strength ($\lambda$) of NoisyTwins loss.}
    \label{fig:CIFAR10LT_Ablation}
    \vspace{-6.0mm}
\end{figure}

\begin{table}[!t]
    \centering
    \caption{\textbf{Quantitative results on ImageNet Carnivore and AnimalFace Datasets.} Our method improves over both StyleGAN2-ADA (SG2-ADA) baseline and SotA Transitional-cGAN .}
    \label{tab:few_shot}
    \resizebox{0.5\textwidth}{!}
    {
    \begin{tabular}{lcc|cc}
    \toprule
        & \multicolumn{2}{c|}{ImageNet Carnivore} & \multicolumn{2}{c}{AnimalFace} \\ \hline
         Method & FID($\downarrow$) & iFID$_\mathrm{CLIP}$($\downarrow$) & FID($\downarrow$) & iFID$_\mathrm{CLIP}$($\downarrow$) \\ \midrule
         SG2~\cite{Karras2019stylegan2}& 111.83 & 36.34 & 94.09 & 29.94 \\
         SG2+ADA~\cite{Karras2020ada}& 22.77 & 12.85 & 20.25 & 11.12 \\
         \midrule
         
        \rowcolor{gray!10}  SG2+ADA+Noise (Ours) & \underline{19.25} & \underline{12.51} & \underline{18.78} & \underline{10.42}\\
        \rowcolor{gray!10} \; + NoisyTwins (Ours) & \textbf{16.01} & \textbf{12.41} & \textbf{17.27} &  \textbf{10.03}\\ \midrule
         & \multicolumn{2}{c|}{FID($\downarrow$)}  & \multicolumn{2}{c}{FID($\downarrow$)} \\  \midrule Transitional-cGAN~\cite{shahbazi2022collapse} & \multicolumn{2}{c|}{14.60} & \multicolumn{2}{c}{20.53} \\
          \rowcolor{gray!10} \; + NoisyTwins (Ours)& \multicolumn{2}{c|}{\textbf{13.65}} & \multicolumn{2}{c}{\textbf{16.15}} \\
        \bottomrule
    \end{tabular}
    }
    \vspace{-5.0mm}
\end{table}
\vspace{1mm} \noindent \textbf{How much noise and regularization strength is optimal?} In Fig. \ref{fig:CIFAR10LT_Ablation}, we ablate over the noise variance parameter $\sigma$ for CIFAR10-LT. We find that a moderate value of noise strength 0.75 leads to optimal results. For the strength of NoisyTwins loss ($\lambda$), we find that the algorithm performs similarly on values near 0.01 and is robust to it (Fig. \ref{fig:CIFAR10LT_Ablation}).

\vspace{1mm} \noindent   \textbf{Which type of self-supervision to use with noise augmentation?}
The goal of our method is to achieve invariance to Noise Augmentation in the $\mc{W}$ latent space. This can be achieved using either contrastive learning-based methods like SimCLR\cite{chen2020simple} or negative-free method like Barlow Twins\cite{zbontar2021barlow}. Contrastive loss (SimCLR based) produces FID of 26.23 vs 17.74 by NoisyTwins (BarlowTwins based). We find that contrastive baseline improves over the noise augmentation baseline (28.90) however falls significantly below the NoisyTwins, as the former requires a large batch size to be effective which is expensive for GANs. 

 \vspace{1mm}\noindent  \textbf{How does NoisyTwins compare with modern Vision and Language models?} For evaluating the effectiveness of modern vision language-based diffusion models, we test the generation of the iNaturalist 2019 dataset by creating the prompt ``a photo of \texttt{S}" where we replace the class name in place of \texttt{S}. We use the LDM~\cite{rombach2021highresolution} model trained on LAION-400M to perform inference, generating 50 images per class. We obtained an FID of 57.04 in comparison to best FID of 11.46 achieved by NoisyTwins. This clearly demonstrates that for specific use cases like fine-grained generation, GANs are still ahead of general-purpose LDM.  
 
\section{Conclusion}
\vspace{-1mm}
In this work, we analyze the performance of StyleGAN2 models on the real-world long-tailed datasets  including iNaturalist 2019 and ImageNet-LT. We find that existing works lead to either class confusion or mode collapse in the image space. This phenomenon is rooted in collapse and confusion in the latent $\mc{W}$ space of StyleGAN2. Through our analysis, we deduce that this collapse occurs when the latents become invariant to random conditioning vectors $\mb{z}$, and collapse for each class. To mitigate this, we introduce inexpensive noise based augmentation for discrete class embeddings. Further, to ensure class consistency, we couple this augmentation technique with BarlowTwins' objective in the latent $\mc{W}$ space which imparts intra-class diversity to latent $\mb{w}$ vectors. The noise augmentation and regularization comprises our proposed NoisyTwins technique, which improves the performance of StyleGAN2 establishing a new SotA on iNaturalist 2019 and ImageNet-LT. The extension of NoisyTwins for conditioning on more-complex attributes for StyleGANs is a good direction for future work. \\
\textbf{Acknowledgements}: This work was supported in part by  SERB-STAR Project (STR/2020/000128). Harsh Rangwani is supported by PMRF fellowship.

\label{sec:intro}

\renewcommand \thepart{}
\renewcommand \partname{}

\doparttoc %
\faketableofcontents %

\appendix
\renewcommand*\contentsname{Appendix}
\addcontentsline{toc}{section}{Appendix} %
\part{Appendix} %
\parttoc %
\begin{figure}[h]
    \centering
    \includegraphics[width=\linewidth]{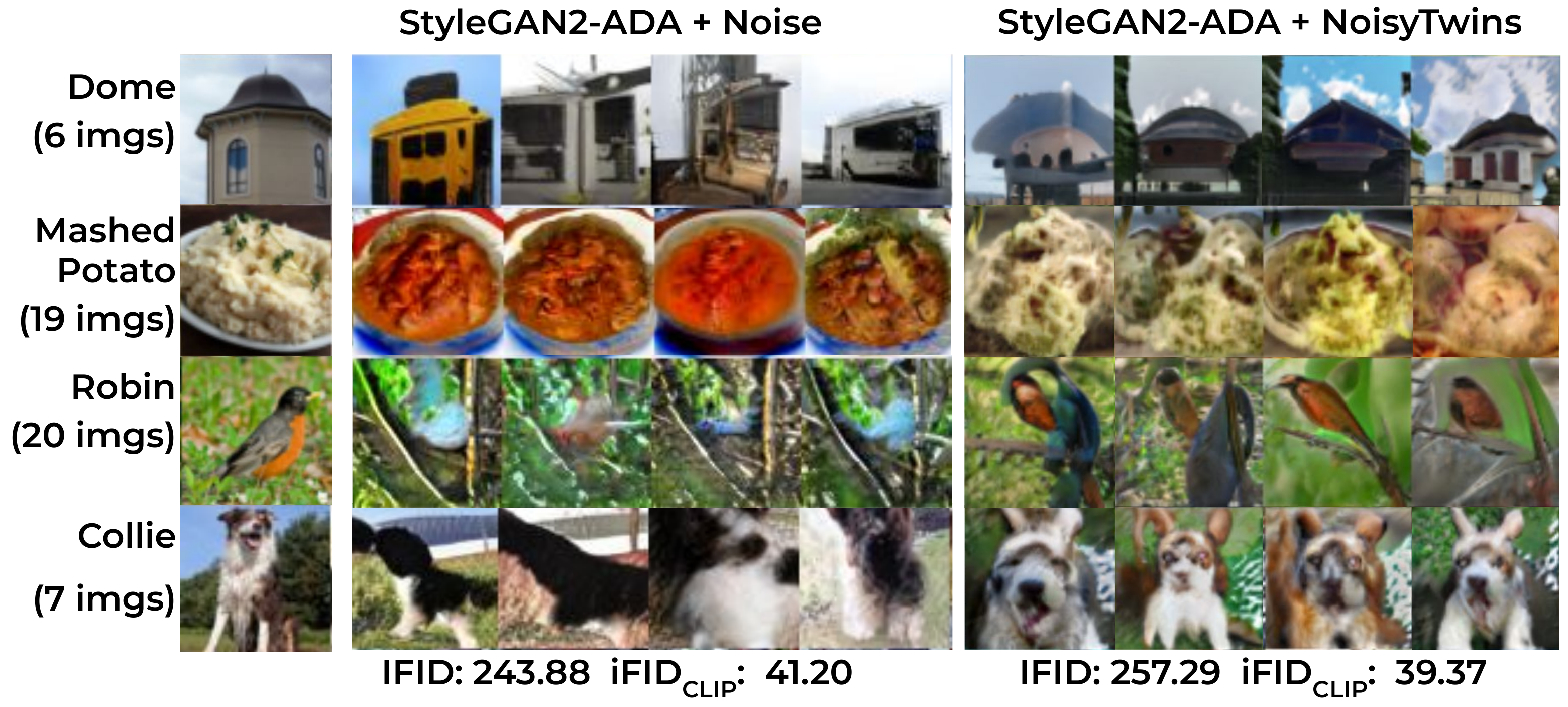}
    \caption{\textbf{Qualitative Results and iFID.} We observe that the noise-only baseline suffers from the mode collapse and class confusion for tail categories as shown on (\emph{left}). Despite this, it is found that the mean iFID based on Inception V3 shows a smaller value for StyleGAN2ADA+Noise, whereas a higher value for diverse and class-consistent NoisyTwins. Hence, this metric does not align with qualitative results. On the other hand, the proposed mean iFID$_\mathrm{CLIP}$ is lower for NoisyTwins, demonstrating its reliability for evaluating GAN models.}
    \label{fig:mean_ifid_imagenet}
\end{figure}

\section{Notations and Code}
We summarize the notations used throughout the paper in Table \ref{tab:notations}. We provide PyTorch-style pseudo code for NoisyTwins in \texttt{noisy\_twins.py} in the supplementary material. We will open-source our code to promote reproducible research.

\section{Comparison of iFID and \texorpdfstring{iFID$_\mathrm{CLIP}$}{iFID-CLIP}}

\begin{figure*}[t]
    \centering
    \includegraphics[width=\linewidth]{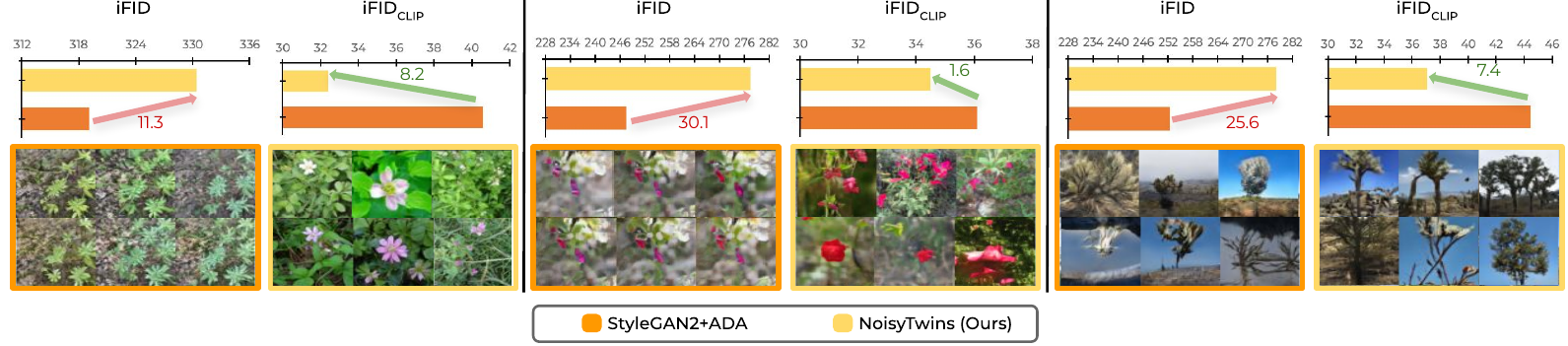}
    \caption{\textbf{iFID Comparison on iNaturalist 2019 dataset.} We provide examples of classes where the quality of images generated by StyleGAN2-ADA is worse, which either suffers from mode collapse or artifacts in generation. Yet iFID based on Inception V3 ranks it higher in terms of quality, which doesn't align with human judgement. On the other hand the proposed iFID$_{\mathrm{CLIP}}$ is able to rank the models correctly and gives a lower value to diverse generations from NoisyTwins.}
    \label{fig:iFID Comparison_3_class}
\end{figure*}

In this section, we present failure cases of InceptionV3-based iFID in the detection of mode collapse, and show how CLIP-based iFID can detect these cases. InceptionV3-based iFID assigns a lower value to a generator with mode collapse, compared to another generator which creates diverse and class-consistent images. In addition to the example given in the main text (Fig. \textcolor{red}{5}), we provide examples from three different classes (Fig.~\ref{fig:iFID Comparison_3_class}). In all the four cases, the InceptionV3-based iFID is better for mode collapsed classes. \emph{Whereas  iFID$_\mathrm{CLIP}$ follows the correct behavior, where the class consistent and diverse model is ranked better}. Due to this inconsistent behavior, mean iFID (mean across classes) which is a commonly used as a metric for quantifying class confusion~\cite{kang2021ReACGAN} can be incorrect. %

 For example, we observe that the StyleGAN2-ADA baseline with proposed noise augmentation achieves mean iFID (243.88) on ImageNet-LT, compared to 257.29 for the NoisyTwins model (Table \textcolor{red}{1} in main text). However, while examining the tail class samples (Fig. \ref{fig:mean_ifid_imagenet}), we find that noise augmented baseline suffers from mode collapse and class confusion, whereas NoisyTwins generates diverse and class-consistent images. Hence, the mean iFID based on Inception-V3 does not align well with qualitative results. On the contrary, the iFID$_\mathrm{CLIP}$ value is 41.20 for the noise-augmented model compared to 39.37 for NoisyTwins, which correlates with the human observation that the NoisyTwins model should have a lower FID as it is diverse and class-consistent. Hence, the proposed metric iFID$_\mathrm{CLIP}$ can be used to to evaluate models for class-conditional image generation reliably.

\begin{table}[!t]
\centering
\caption{\textbf{Notation Table}}
\label{tab:notations}
\resizebox{\linewidth}{!}{%
\begin{tabular}{p{0.17\linewidth} p{0.15
\linewidth} p{0.68\linewidth}}
\toprule
Symbol                & Space                              & Meaning                                                                               \\ \midrule
$\mb{c}$                     & $\mathbb{R}^{d}$                    & Class Embedding                                                                           \\
$\mb{z}$            & $\mathbb{R}^{d}$                          & Noise vector                                                                          \\
$\mathbf{w}$            & $\mathbb{R}^{d}$                   & Vector in $\mc{W}$ latent Space   \\
$\mc{D}$                     &                                    & Discriminator    \\
$\mc{G}$                     &                                    & Generator      \\
$BS$                    &  $\mathbb{R}^+$                      & Batch Size        \\
$\mathbf{x_i}$          & $\mathbb{R}^{3 \times H \times W}$ & Image  \\
$\mb{\tilde{c}}$          & $\mathbb{R}^{d}$                   & Noise Augmented Class Embedding     \\
$n_c$                    & $\mathbb{R}^{+}$                   & Frequency of training samples in class $\mb{c}$      \\
$\sigma_c$                & $\mathbb{R}^{+}$                   & Effective number of samples based noise standard deviation     \\
$\sigma$                & $\mathbb{R}^+$                        & Hyperparameter for scaling noise      \\
$\mb{\mu_{c}}$      & $\mathbb{R}^d$                            & Mean embedding parameters of class $\mb{c}$      \\
$\tilde{\mb{W}}_A$ $\tilde{\mb{W}}_B$ & $\mathbb{R}^{BS \times d} $          & Batches of augmented latents      \\
$\mb{C}_{j,k}$     &    $\mathbb{R}$                            & Cross-correlation between $j$th and $k$th latent variables        \\
$\lambda$          & $\mathbb{R}^{+}$                           & Strength of NoisyTwins regularization    \\
$\gamma$            & $\mathbb{R}^{+}$                          & Relative importance of the two terms of NoisyTwins loss     \\
$\rho$                  & $\mathbb{R}^{+}$                       & Imbalance ratio of dataset: Ratio between the most and the least frequent classes\\ \bottomrule
\end{tabular}}
\end{table}

\section{Experimental Details}

\begin{table*}[!h]
    \parbox{\textwidth}{
        \centering
        \caption{\textbf{HyperParameter Configurations used for experiments.} We provide a detailed list of hyperparameters used for the experiments across datasets for NoisyTwins on StyleGANs.}
        \label{tab:hyperparameters}}
        \resizebox{\textwidth}{!}
        {
        \begin{tabular}{r|c|c|c||c|c}
        \toprule
            \multicolumn{1}{c}{}& \multicolumn{3}{c}{Long-Tail Datasets} & \multicolumn{2}{c}{Few-Shot Datasets} \\
            \midrule
              & iNaturalist-2019 & ImageNet-LT & CIFAR10-LT ($\rho$=100) & ImageNet Carnivores & AnimalFaces  \\
            \midrule
            Resolution & 64 & 64 & 32 & 64 & 64 \\
            Augmentation & ADA & ADA & DiffAug & ADA & ADA \\
            \midrule 
             \multicolumn{1}{c}{}& \multicolumn{5}{c}{Regularizers}\\
            \midrule
            Effective Samples $\alpha$ & 0 & 0 & 0.99 & 0 & 0  \\
            Noise Scaling $\sigma$ & 0.1 & 0.25 & 0.75 & 0.5 & 0.5 \\
            NoisyTwins Start Iter. & 25k & 60k & 0 &  0 & 0 \\
            NoisyTwins Weights ($\lambda$, $\gamma$) & 0.001, 0.005 & 0.001, 0.005 & 0.01, 0.05 & 0.001, 0.05 & 0.001, 0.05 \\
            LeCam Reg Weight & 0.01 & 0 & 0 & 0 & 0 \\ 
            R1 Regularization $\gamma_{R1}$ & 0.2048 & 0.2048 & 0.01 & 0.01 & 0.01 \\
            PLR Start Iter. & 0 & 60k & No PLR & 0 & 0 \\
            \midrule 
            \multicolumn{1}{c}{} & \multicolumn{5}{c}{StyleGAN}\\
            \midrule
            Mapping Net Layers & 2 & 8 & 8 & 2 & 2 \\
            $\mc{D}$ Backbone & ResNet & ResNet & Orig & ResNet & ResNet \\
            Style Mixing & 0.9 & 0.9 & 0 & 0 & 0\\
            $\mc{G}$ EMA Rampup & None & None & 0.05 & 0.05 & 0.05\\
            $\mc{G}$ EMA Kimg & 20 & 20 & 500 & 500 & 500\\
            MiniBatch Group & 8 & 8 & 32 & 32 & 32 \\
            
            \bottomrule
        \end{tabular}
        }
\end{table*}
\newpage

\begin{table*}[!t]
    \centering
    \parbox{\textwidth}{
    \caption{
    \textbf{Statistical Analysis for CIFAR10-LT.} This table provides the mean and one standard deviation of metrics for all methods on CIFAR10-LT performed on three independent evaluation runs by generating 50k samples across random seeds. 
    }
        \label{tab:CIFAR10_LT Std Table}}
    \resizebox{0.9\textwidth}{!}{
    \begin{tabular}{lccccc}
    \toprule
         & \multicolumn{5}{c}{CIFAR10-LT ($\rho$=100)} \\ \hline
         Method & FID($\downarrow$) & FID$_\mathrm{CLIP}$($\downarrow$) & iFID$_\mathrm{CLIP}$($\downarrow$) & Precision($\uparrow$) & Recall($\uparrow$)\\
         \midrule
         SG2+DiffAug~\cite{Karras2020ada}& 31.72$_{\pm{0.16}}$ & 6.24$_{\pm{0.02}}$ & 11.63$_{\pm{0.03}}$ & 0.63$_{\pm{0.00}}$ & 0.35$_{\pm{0.00}}$ \\
         SG2+D2D-CE~\cite{kang2021ReACGAN}& 20.08$_{\pm{0.15}}$ & 4.75$_{\pm{0.04}}$ & 11.35$_{\pm{0.01}}$ & \textbf{0.73}$_{\pm{0.00}}$ & 0.43$_{\pm{0.00}}$ \\
         gSR~\cite{rangwani2022gsr}& 22.50$_{\pm{0.29}}$ & 5.55$_{\pm{0.01}}$ & 9.94$_{\pm{0.00}}$ & 0.70$_{\pm{0.00}}$ & 0.28$_{\pm{0.01}}$ \\
         \midrule
         
        \rowcolor{gray!10}  SG2+DiffAug+Noise (Ours)& 28.85$_{\pm{0.18}}$ & 5.29$_{\pm{0.02}}$ & 10.64$_{\pm{0.01}}$ & 0.71$_{\pm{0.00}}$ & 0.38$_{\pm{0.00}}$ \\
        \rowcolor{gray!10} \; + NoisyTwins (Ours)& \textbf{17.72}$_{\pm{0.08}}$ & \textbf{3.56}$_{\pm{0.01}}$ & \textbf{7.27}$_{\pm{0.02}}$ & 0.69$_{\pm{0.01}}$ & \textbf{0.52} $_{\pm{0.01}}$ \\ \bottomrule
    \end{tabular}
    }
    
\end{table*}

\begin{figure}[!t]
    \centering
    \includegraphics[width=\linewidth]{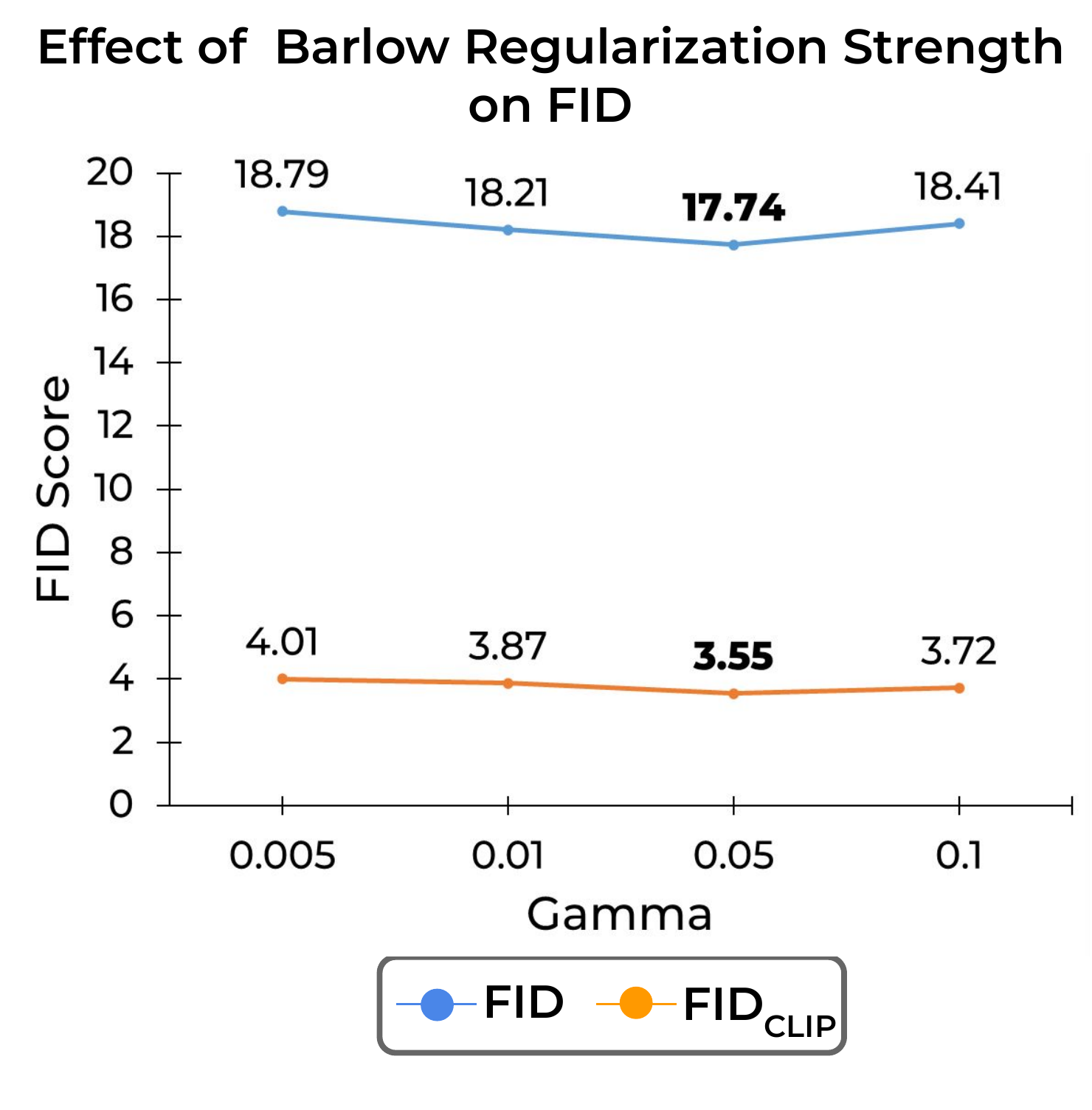}
    \caption{\textbf{Ablation on \texorpdfstring{$\gamma$}{Gamma}:} Quantitative comparison on CIFAR10-LT for the strength of hyperparameter (\texorpdfstring{$\gamma$}{Gamma}) in NoisyTwins loss function.}
    \label{fig:Barlow Momentum Ablation}
\end{figure}

We run our experiments using PyTorchStudioGAN~\cite{kang2022StudioGAN} as the base framework. For most baseline experiments, we use the standard StyleGAN configurations present in the framework. We use a learning rate of 0.0025 for the discriminator ($\mc{D}$) and the generator ($\mc{G}$) network. We use a batch size of 128 for all our experiments. In addition, following the observations of previous work~\cite{Sauer2021ARXIV}, we apply a delayed Path Length Regularization (PLR) starting at 60k iterations for all our experiments on ImageNet-LT. For NoisyTwins, the most important hyperparameters are $\lambda$ (regularization strength) and $\sigma$ (noise variance). We perform a grid search on $\lambda$ values of \{0, 0.001, 0.01, 0.1\} and $\sigma$ values of \{0.10, 0.25, 0.50, 0.75\}.  We provide a detailed list of optimal hyperparameters used in Table \ref{tab:hyperparameters}. All the models trained on a particular dataset use the same hyperparameters, to maintain fairness in the comparison of models. We summarize all the hyperparameters used for respective datasets in Table~\ref{tab:hyperparameters}.%

For our experiments on few-shot datasets with SotA transitional-cGAN, we use the author's official code implementation available on GitHub~\footnote{https://github.com/mshahbazi72/transitional-cGAN}. We use the same configuration specified to first evaluate on ImageNet Carnivores and AnimalFaces datasets. To integrate NoisyTwins, we generate the noise augmentations by augmenting the class embeddings and then apply NoisyTwins regularization in $\mc{W}$ space. We use the same hyperparameter setting used by the authors and NoisyTwins with $\lambda = 0.001$ and $\gamma = 0.05$.

\subsection{Statistical Significance of the Experiments}

We report mean and standard deviation over three evaluation runs for all baselines on the CIFAR10-LT (Table~\ref{tab:CIFAR10_LT Std Table}). It can be observed that most metrics that we have reported have a low standard deviation, and metrics are close to the mean value across runs. As we find standard deviation to be low across the metrics evaluated and the process of evaluating iFID to be expensive, we do not explicitly report them on large multi-class datasets. 

\begin{table*}[!t]
    \centering
    \parbox{\textwidth}{
    \caption{ \textbf{Evaluation of NoisyTwins by varying degree of imbalance.} NoisyTwins can produce diverse and class-consistent results across imbalance ratios. 
    }
        \label{tab:CIFAR10-LT imb factor ablation}}
    \resizebox{0.9\textwidth}{!}{
    \begin{tabular}{lcccccc}
    \toprule
         & \multicolumn{5}{c}{CIFAR10-LT} \\ \hline
         Method & $\rho$ & FID($\downarrow$) & FID$_\mathrm{CLIP}$($\downarrow$) & iFID$_\mathrm{CLIP}$($\downarrow$) & Precision($\uparrow$) & Recall($\uparrow$)\\
         \midrule
         SG2+DiffAug~\cite{Karras2020ada}& \multirow{2}{*}{50} & 26.79 & 5.83 & 9.61 & 0.65 & 0.38\\
        \;+NoisyTwins (Ours) & & \textbf{14.92} & \textbf{2.99} & \textbf{6.38} & \textbf{0.71} & \textbf{0.57}\\ 
        \midrule
        SG2+DiffAug~\cite{Karras2020ada}& \multirow{2}{*}{100} & 31.73 & 6.27 & 11.59 & 0.63 & 0.35  \\
        \;+NoisyTwins (Ours)& & \textbf{17.74} & \textbf{3.55} & \textbf{7.24} & \textbf{0.70} & \textbf{0.51}\\ 
        \midrule
        SG2+DiffAug~\cite{Karras2020ada}& \multirow{2}{*}{200} & 55.48 & 10.59 & 19.49 & 0.65 & 0.36 \\
        \;+NoisyTwins (Ours)& & \textbf{23.57} & \textbf{4.91} & \textbf{9.17} & \textbf{0.68} & \textbf{0.46}\\ 
        \bottomrule
    \end{tabular}
    }
    
\end{table*}

\begin{figure}[!t]
    \centering
    \includegraphics[width=\linewidth]{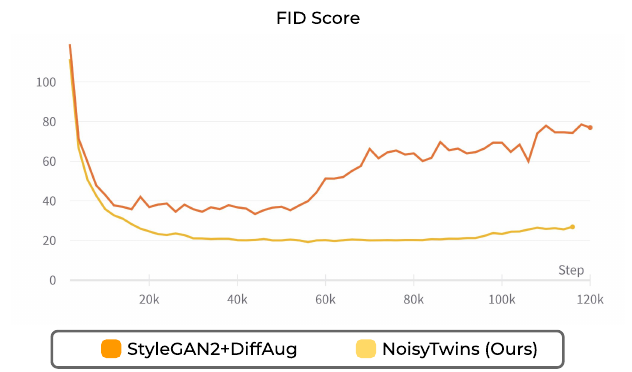}
    \caption{\textbf{Comparison of FID curves for CIFAR10-LT ({\boldmath$\rho$}=100).} NoisyTwins leads to stable training with decreasing FID with iterations.}
    \label{fig:FID Comparison}
\end{figure}

\section{Additional Details of Analysis}
We perform our ablation experiments on CIFAR10-LT using the same configuration as mentioned in Table~\ref{tab:hyperparameters}. We provide ablation experiments on the standard deviation of noise ($\sigma$) and the strength of regularization loss ($\lambda$) (Sec. \textcolor{red}{6}), as we observe that they influence the performance of the system most. We further provide ablation on the parameter $\gamma$ in Fig. \ref{fig:Barlow Momentum Ablation}, which controls the relative importance between the invariance enforcement and decorrelation enhancement terms in Eq. \textcolor{red}{6} of the main text. We find that performance remains almost the same while varying $\gamma$ from 0.005 to 0.1, with optimal value occurring around 0.05 for CIFAR10-LT. Hence, the model is robust to $\gamma$. 

We further analyze our method for a range of imbalance ratios (i.e., $\rho$, ratio of the most frequent to least frequent class) in the class distribution. We present results for CIFAR10-LT with imbalance factors ($\rho$) values of 50, 100, and 200 in Table \ref{tab:CIFAR10-LT imb factor ablation}. Our method can prevent mode collapse and improves the baseline FID significantly in all cases. Also note that the baseline gets more unstable (high FID) as the imbalance ratio increases, which shows the necessity of using NoisyTwins as it stabilizes the training even when large imbalances are present in the dataset (Fig. \ref{fig:FID Comparison}).

\vspace{1mm} \noindent \textbf{Information Maximization in $\mb{w}$:}
Noisy Twins works on the principle of information maximization (IM), same as BarlowTwins~\cite{zbontar2021barlow} (App. Sec. \textcolor{red}{A}), where we maximize the information $\mathcal{I}$ between mapping $\mathbf{w}$ and the inputs $[\mathbf{c},\mathbf{z}]$ to the GAN. This ensures variations in $\mathbf{z}$ are preserved when transformed to $\mathbf{w}$ vector in $\mathcal{W}$-space. To verify this hypothesis, we put the ($\gamma=0$) for IM (cross-correlation) term in Eq.~\ref{eq:noise_aug_eq}, which leads to mode collapse (FID 80). 

\section{Additional Results}
\begin{figure}[!t]
    \centering
    \includegraphics[width=\linewidth]{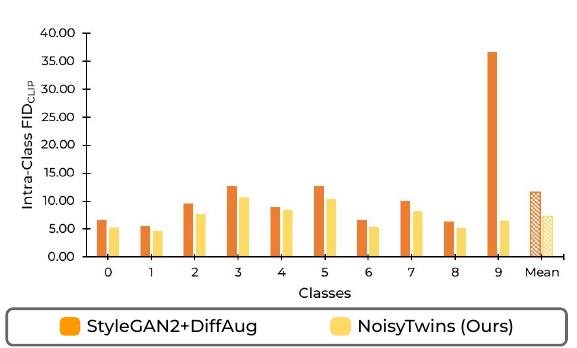}
    \caption{\textbf{Class-wise iFID$_{\textbf{CLIP}}$} comparison of models on CIFAR10-LT (\boldmath$\rho$=100) dataset.}
    \label{fig:iFID Class-Wise Comparison}
\end{figure}

Fig. \ref{fig:iFID Class-Wise Comparison}  provides the class-wise comparison of the proposed iFID$_\mathrm{CLIP}$ for the baseline and after adding NoisyTwins. NoisyTwins produces better iFID$_\mathrm{CLIP}$ for all classes, hence does not lead to performance degradation for head classes while improving performance on tail classes.
      
\begin{figure*}[!t]
    \centering
    \includegraphics[width=\linewidth]{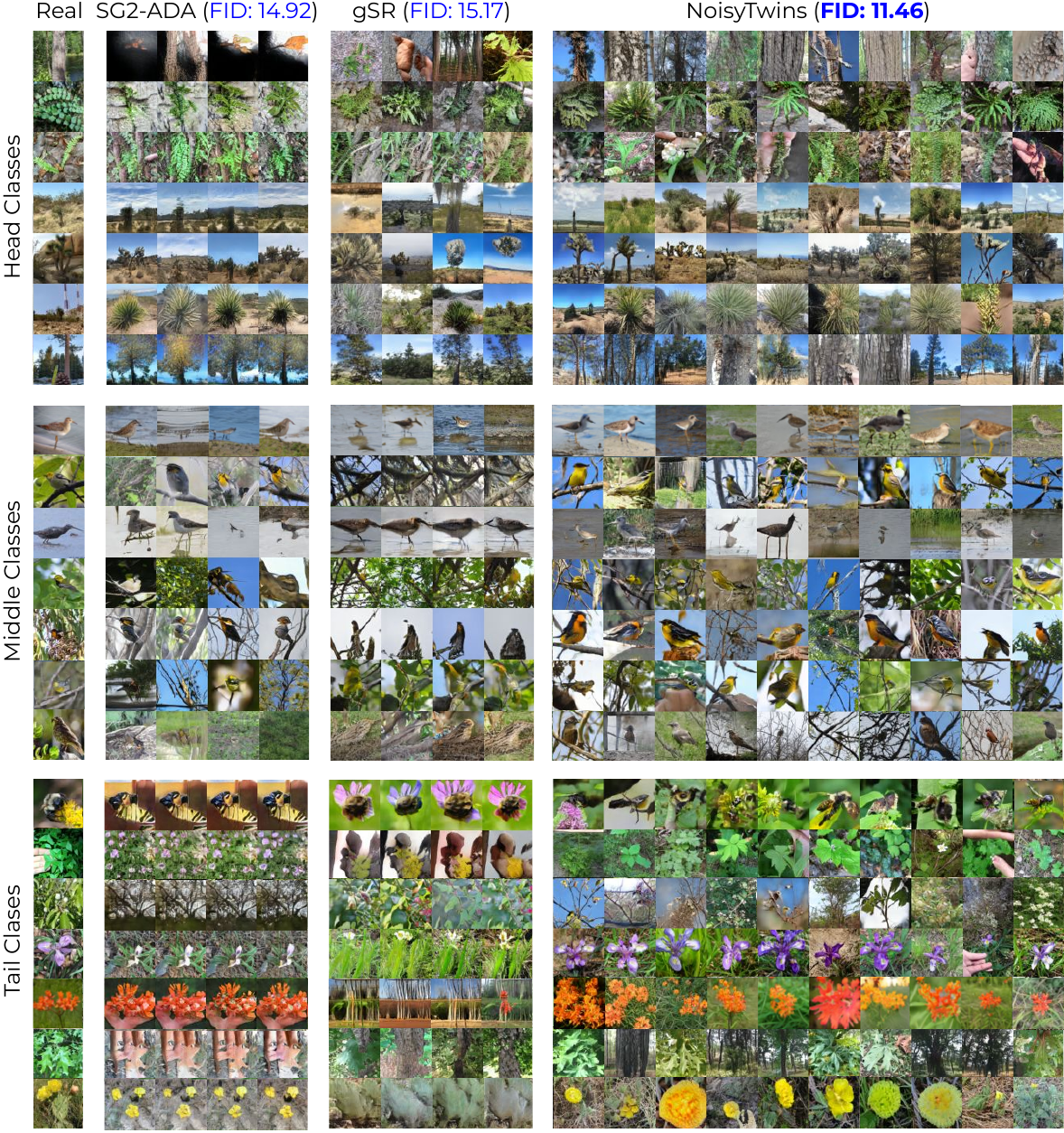}
    \caption{\textbf{Qualitative Analysis on iNaturalist2019 (1010 classes).} Examples of generations from various classes for evaluated baselines (Table \textcolor{red}{1}). The baseline ADA suffers from mode collapse, whereas gSR suffers from class confusion particularly for tail classes, particularly for tail classes as seen above on the left. NoisyTwins  generates diverse and class-consistent images across all categories.}
    \label{fig:iNat_qualitative}
    \vspace{20mm}
\end{figure*}

We now provide additional qualitative results for models. Similar to ImageNet-LT, we also provide a full-scale comparison of images from different methods in Fig. \ref{fig:iNat_qualitative} for iNaturalist-2019. In addition to the images from the tail classes, we also show generations from the head and middle classes. In Fig. \ref{fig:iNat_qualitative}, it is clearly shown that NoisyTwins can obtain high-quality and diverse samples compared to the baseline. We find that the StyleGAN2-ADA baseline produces similar images across a class for tail classes, which confirms the occurrence of class-wise mode collapse even in large datasets. Further, it can be seen that the regularizer-based method (gSR) is unable to capture the identity of the real class and suffers from the issue of class confusion (as also seen in t-SNE of Fig. \textcolor{red}{2} of the main text). Our method NoisyTwins, can produce realistic-looking diverse images even for tail classes, which shows the successful transfer of knowledge from head classes. Training a class-conditioned GAN on long-tailed datasets leads to class confusion when the extent of knowledge transfer is not controlled. NoisyTwins strikes the right balance between knowledge transfer from the head classes to benefit the quality of generation in the tail classes, thus not allowing class confusion. This would not be possible if we train GAN independently on tail classes ($\sim$ 30 images), which shows the practical usefulness of joint training on complete long-tailed data (i.e., our setup).

\begin{figure*}[h]
    \centering
    \includegraphics[width=\linewidth]{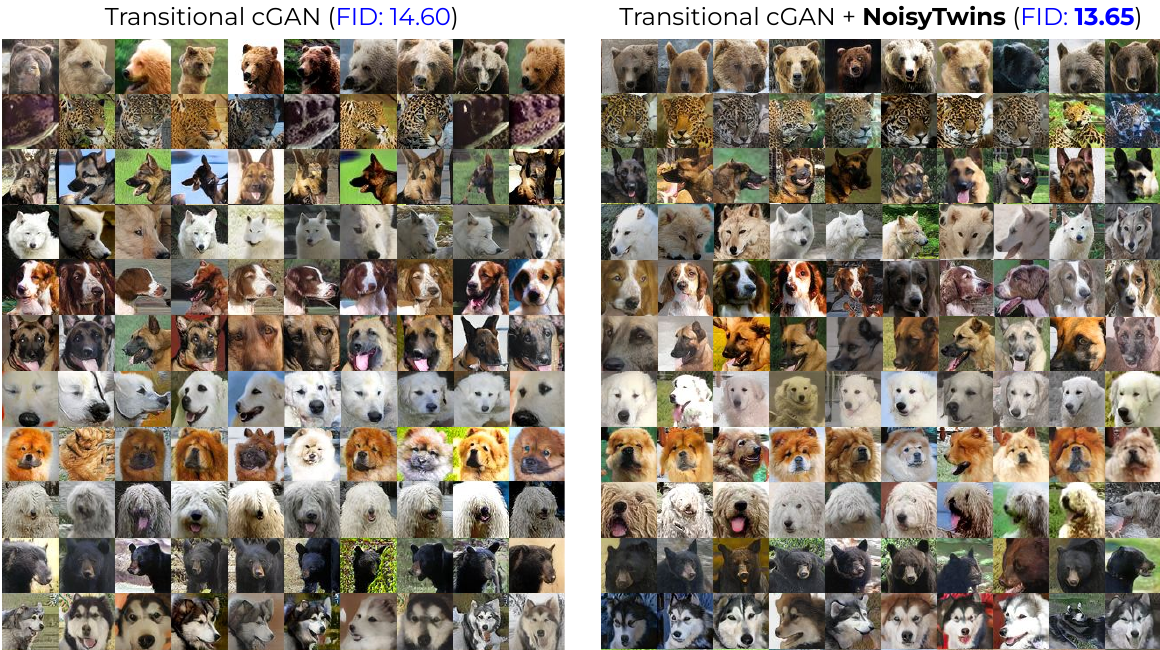}
    \caption{\textbf{Qualitative comparison on few-shot ImageNet Carnivores dataset.}}
    \label{fig:in_carnivore}
\end{figure*}
\begin{figure*}[h]
    \centering
    \includegraphics[width=\linewidth]{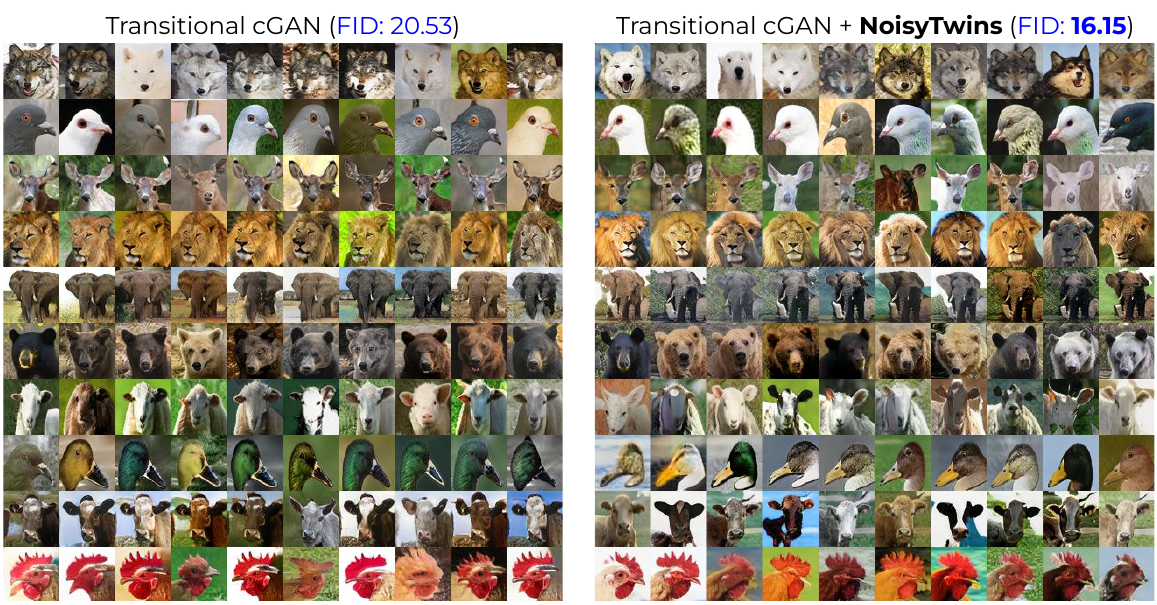}
    \caption{\textbf{Qualitative comparison on few-shot AnimalFaces dataset.}}
    \label{fig:animalface}
    \vspace{5mm}
\end{figure*}

We showcase qualitative results of generations from few-shot datasets (i.e., ImageNet Carnivore and AnimalFaces). Fig. \ref{fig:in_carnivore} and \ref{fig:animalface} show the results of the SotA few-shot baseline of Transitional-cGAN  (\textit{left}) and after augmenting it with our proposed NoisyTwins (\textit{right}). Our proposed method, NoisyTwins, can further stabilize the training of Transitional-cGAN and improve the quality and diversity of the generated samples on both datasets of ImageNet Carnivores and AnimalFaces.

\begin{table}[!t]
    \centering
   
    \caption{Results for Large Resolutions on Animal Faces dataset}
     \label{tab:AF-high-res}

    \begin{adjustbox}{max width=\columnwidth}
    \begin{tabular}{l|ccc}
        \hline
     { FID ($\downarrow$)} & AF (128 $\times$ 128) & AF (256 $\times$ 256) \\ %
    \hline

    \textbf{Transitional-cGAN}~\cite{shahbazi2022collapse}  & 22.59 & 22.28 \\  %
    
     \textbf{+NoisyTwins (Ours)}& \textbf{16.79} & \textbf{19.14} \\ %

     \hline

    \end{tabular}

\end{adjustbox}
\end{table}

\begin{table}[!t]
    \centering

     \caption{Results for large iNaturalist 2019 dataset (128 $\times$ 128)}
     \label{tab:inat-128}
    \begin{adjustbox}{width=\columnwidth}
    \begin{tabular}{l|ccc}
    \hline
         &  FID (80k) ($\downarrow$) & FID ($\downarrow$) & FID$_\mathrm{CLIP}$ ($\downarrow$)  \\ %
    \hline
    \textbf{StyleGAN2-ADA}~\cite{Karras2020ada}  & 16.58 & 12.31 & 2.18 \\  %
    
     \textbf{+NoisyTwins (Ours)} & \textbf{15.29} & \textbf{12.01} &  \textbf{1.93}\\ %
     \hline 
    
    \end{tabular}
 \end{adjustbox}
\vspace{10mm}
\end{table}
\vspace{1mm} \noindent \textbf{Results across other Resolutions:} NoisyTwins scales well on larger resolutions as demonstrated on few-shot AnimalFaces (AF) dataset using Transitional-cGAN~\cite{shahbazi2022collapse} in Table~\ref{tab:AF-high-res}, where we observe a significant improvement if FID for both $128 \times 128$ and $64 \times 64$ resolution data. Further, on large-scale iNat-19 StyleGAN2-ADA baseline in Tab. ~\ref{tab:inat-128}, we also find that NoisyTwins is able to improve performance. The NoisyTwins method also converges faster as at intermediate stage of 80k iterations in full run of 150k iterations, the FID for NoisyTwins is lower than baseline.  As NoisyTwins method is based on the information maximization principle~\cite{zbontar2021barlow} and generalizes on datasets, we expect it benefits other large resolutions of StyleGAN too, similar to what is observed in Sauer \etal ~\cite{Sauer2021ARXIV}.

\begin{figure}[!t]
    \centering
    \includegraphics[width=\linewidth]{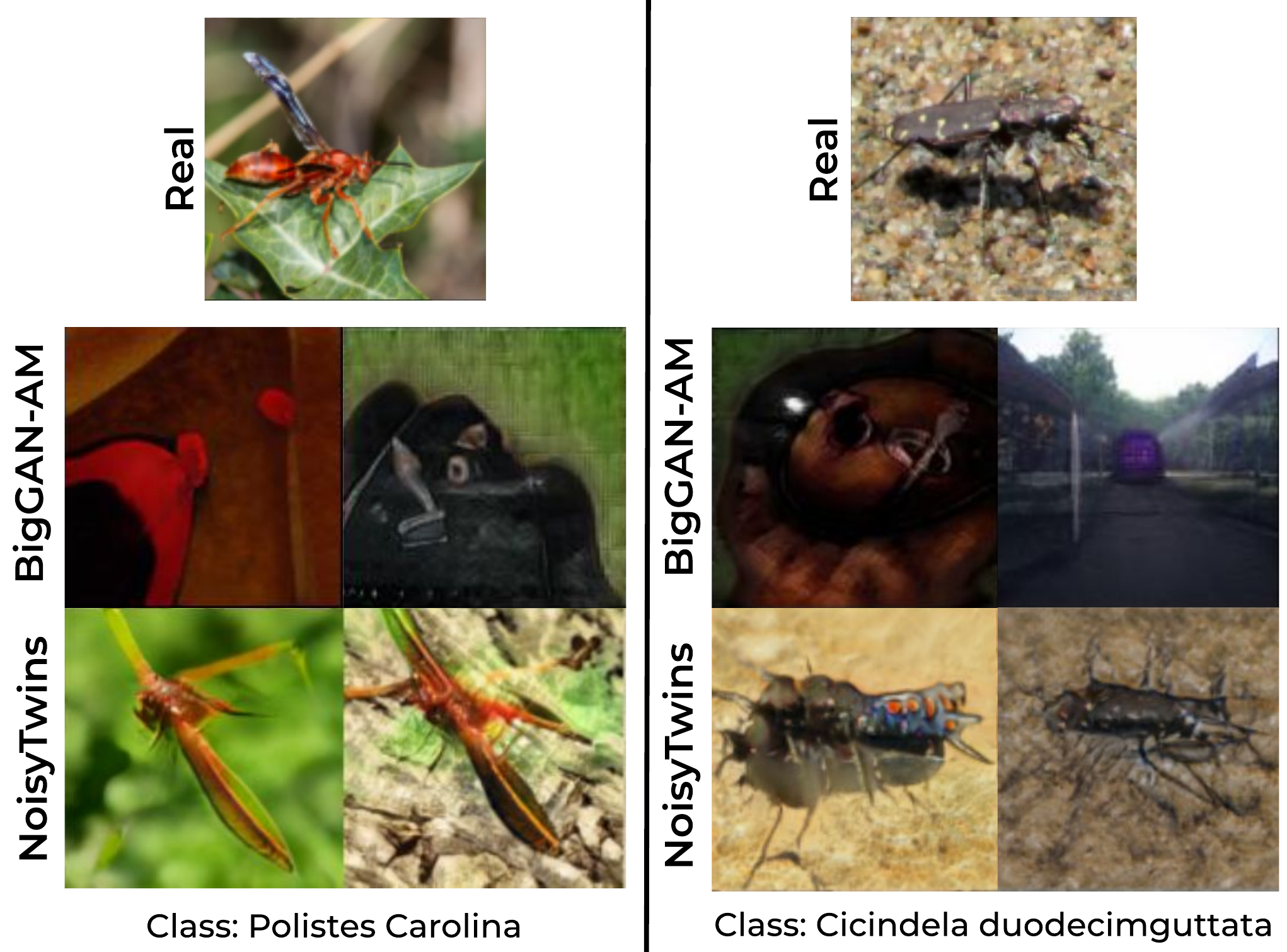}
    \caption{\textbf{BigGAN-AM results on iNaturalist Dataset.}}
    \label{fig:finetuning_fig}
\end{figure}

\begin{figure}[!t]
    \includegraphics[width=\linewidth]{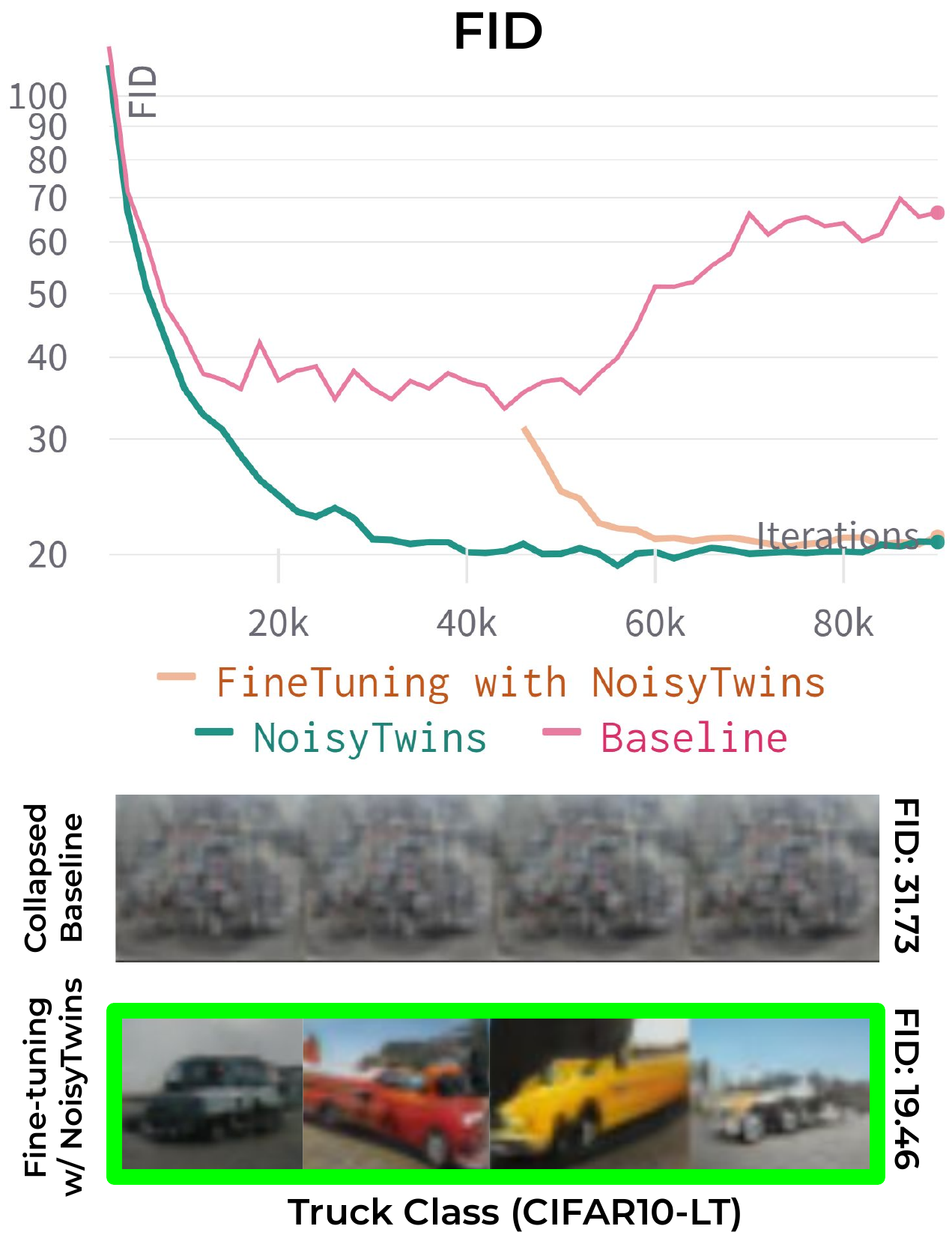}
    \caption{\textbf{Fine-tuning Results.} (Top) FID Curve during fine-tuning with NoisyTwins for CIFAR10-LT dataset. (Below) Diverse images of the truck class generated after fine-tuning baseline with NoisyTwins.}
    \label{fig:finetuneFID_fig}
\end{figure}

\vspace{1mm} \noindent \textbf{Comparison to Fine-Tuning Approaches:} We tested NoisyTwins in fine-tuning setting to investigate if it is able to overcome mode collapse. For this we first train StyleGAN-2 DiffAug baseline (Table \textcolor{red}{2}) and then obtain the checkpoint which has collapse, we then resume training of baseline after adding the NoisyTwins regularizer. As seen in Fig. \ref{fig:finetuneFID_fig}, NoisyTwins is able to reconstruct the collapsed class of baseline on fine-tuning, improving the FID to 19.46 from 31.73 on the CIFAR10-LT dataset.

We also compare our method to other fine-tuning approaches like BigGAN-AM~\cite{li2020cost}, which tries to adapt the embeddings for new classes or repair collapsed classes using knowledge transfer from a pre-trained classifier trained on the target dataset. However, we see in Fig. \ref{fig:finetuning_fig}, when fine-tuned for fine-grained datasets like iNaturalist, these approaches fail completely due to the significant domain shift of these datasets compared to ImageNet. We hypothesize that this is because the activation maximization(AM)~\cite{li2020cost} using a classifier trained on iNaturalist is unable to produce meaningful images as there is presence of distribution shift between the datasets.

\clearpage

{\small
\bibliographystyle{ieee_fullname}
\bibliography{egbib}
}

\end{document}